\documentclass[sigconf]{acmart}

\usepackage{multicol}

\AtBeginDocument{%
  \providecommand\BibTeX{{%
    \normalfont B\kern-0.5em{\scshape i\kern-0.25em b}\kern-0.8em\TeX}}}

\copyrightyear{2022}
\acmYear{2022}
\setcopyright{acmcopyright}\acmConference[SIGGRAPH '22 Conference Proceedings]{Special Interest Group on Computer Graphics and Interactive Techniques Conference Proceedings}{August 7--11, 2022}{Vancouver, BC, Canada}
\acmBooktitle{Special Interest Group on Computer Graphics and Interactive Techniques Conference Proceedings (SIGGRAPH '22 Conference Proceedings), August 7--11, 2022, Vancouver, BC, Canada}
\acmPrice{15.00}
\acmDOI{10.1145/3528233.3530755}
\acmISBN{978-1-4503-9337-9/22/08}

\acmSubmissionID{813}

\begin{document}

\title{Single-View View Synthesis in the Wild with Learned Adaptive Multiplane Images}

% \affiliation{%

\author{Yuxuan Han}
\authornote{Work done when YH and RW were interns at MSRA mentored by JY, who is the corresponding author.}
\affiliation{%
  \institution{Tsinghua University}
  \city{Beijing}
  \country{China}
  \orcid{0000-0002-2844-5074}
}
\email{hanyuxuan076@gmail.com}

\author{Ruicheng Wang}
\authornotemark[1]
\affiliation{%
  \institution{USTC}
  \city{Hefei}
  \country{China}
  \orcid{0000-0003-3082-0512}
}
\email{wangrc2018cs@mail.ustc.edu.cn}

\author{Jiaolong Yang}
\affiliation{%
  \institution{Microsoft Research Asia}
  \city{Beijing}
  \country{China}
  \orcid{0000-0002-7314-6567}
}
\email{jiaoyan@microsoft.com}
\renewcommand{\shortauthors}{Han, Wang, and Yang}

%%
%% some command
\newcommand{\hyxtodo}[1]{\textcolor{red}{{[HYX TODO: #1]}}}
\newcommand{\hyxadd}[1]{\textcolor{blue}{{[ADD: #1]}}}
\begin{abstract}
This paper deals with the challenging task of synthesizing novel views for in-the-wild photographs. Existing methods have shown promising results leveraging monocular depth estimation and color inpainting with layered depth representations. However, these methods still have limited capability to handle scenes with complex 3D geometry. We propose a new method based on the multiplane image (MPI) representation. To accommodate diverse scene layouts in the wild and tackle the difficulty in producing high-dimensional MPI contents, we design a network structure that consists of two novel modules, one for plane depth adjustment and another for depth-aware color prediction. The former adjusts the initial plane positions using the RGBD context feature and an attention mechanism. Given adjusted depth values, the latter predicts the color and density for each plane separately with proper inter-plane interactions achieved via a feature masking strategy. To train our method, we construct large-scale stereo training data using only unconstrained single-view image collections by a simple yet effective warp-back strategy.
The experiments on both synthetic and real datasets demonstrate that our trained model works remarkably well and achieves state-of-the-art results.

\end{abstract}




\begin{teaserfigure}
\centering
    \includegraphics[width=1.0\textwidth]{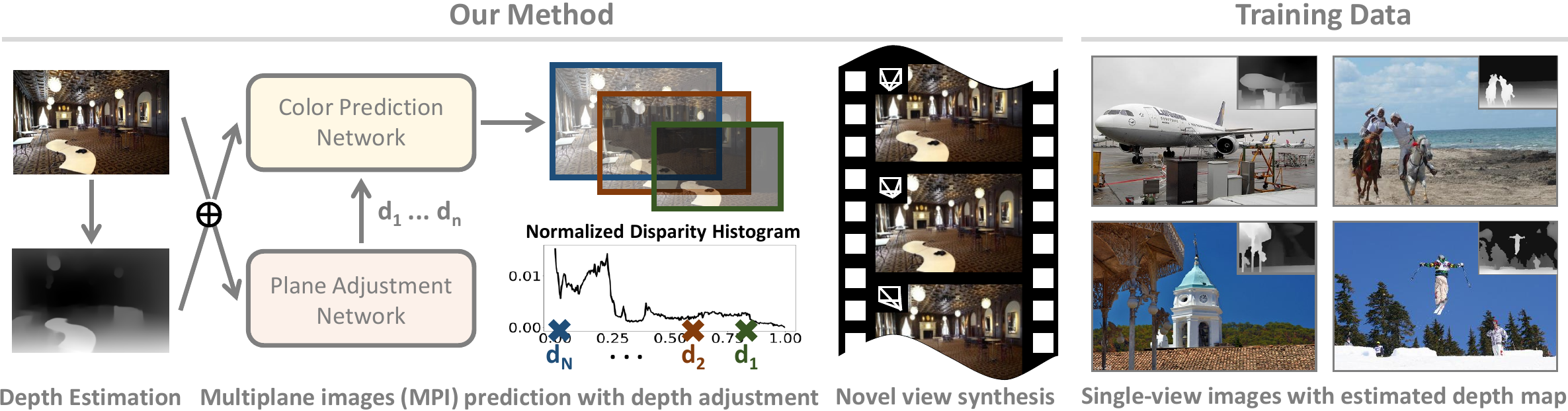}
    \vspace{-18pt}
    \caption{Given a single color image and a depth map estimated by off-the-shelf monocular depth estimators, our method predicts a multiplane image (MPI) with plane depth adjustment for novel view synthesis. Our training dataset is constructed using single-view images in the wild (COCO), as shown on the right. See our \href{https://yxuhan.github.io/AdaMPI}{\emph{project page}} for synthesized videos.
    }
    \label{fig:teaser}
    \vspace{10pt}
\end{teaserfigure}

\maketitle

\section{Introduction}
Learning-based single-view view synthesis has attracted much attention in recent years~\cite{wiles2020synsin,single_view_mpi,Niklaus_TOG_2019}, for it enables 
appealing 3D visual effect 
and facilitates various applications in virtual reality and animation.  
Existing works~\cite{mine2021,rombach2021geometry,hu2021worldsheet,Rockwell2021} have demonstrated promising results 
for specific scene categories such as indoor scene, buildings, and street view in self-driving~\cite{zhou2018stereo,single_view_mpi}. 
However, the problem is still quite challenging for scaling these methods to arbitrary photos in the wild due to the limited power of the scene representation proposed by previous works and the lack of large-scale multi-view image datasets.

A few recent works~\cite{Shih3DP20,jampani:ICCV:2021,Niklaus_TOG_2019,kopf2019practical,kopf2020one} attempted to tackle single-view view synthesis for in the wild images using layered depth representations.
Monocular depth estimation~\cite{Ranftl2020, Ranftl2021} is typically used as the proxy ground-truth depth of the scene to guide novel view generation.
\cite{Shih3DP20} heuristically decompose the scene into a set of layers based on depth discontinuities and train a network to inpaint each occluded layer.
However, it struggles to model thin structures due to its hard depth layering. 
\cite{jampani:ICCV:2021} propose a soft-layer representation to solve the problems in~\cite{Shih3DP20}.
However, their method considers only two layers (foreground and background) and thus is difficult to handle scenes with complex occlusion among multiple objects.

In this work, we adopt multiplane image (MPI)~\cite{zhou2018stereo} as our scene representation which has superb representation power as demonstrated by previous works \cite{zhou2018stereo,srinivasan2019pushingmpi,single_view_mpi,mildenhall2019local,Li2020LF}. 
Most previous methods set the planes in MPI at fixed positions -- typically uniformly-placed in inverse depth or disparity space -- and predict texture by a Convolutional Neural Network (CNN). 
However, MPI is a highly over-parameterized representation~\cite{zhou2018stereo}, which is difficult to learn for neural networks as dozens or even hundreds of channels are required as output. Consequently, the performance may even degrade when the number of planes increases~\cite{Li2020LF}.  
This problem becomes even more pronounced for in-the-wild images as generally more planes are needed to represent diverse scene layouts. 
Some works~\cite{zhou2018stereo,single_view_mpi} propose to reduce the output channels by predicting only one color image and obtain the RGB channel for each plane by blending it with the input image.
Clearly, this strategy sacrifices the representation power of MPI. 
Our key observations to solve this problem are twofold: \emph{i}) the predefined MPI depths are suboptimal for its scene-agnostic nature; \emph{ii}) the network architecture can be carefully designed to mitigate the issue caused by the large output space.

To this end, we present a novel \emph{AdaMPI}  architecture for single-view view synthesis in the wild.
Similar to previous methods designed for in-the-wild scenario, monocular depth estimation~\cite{Ranftl2020,Ranftl2021} is used in our method.
Based on the contextual information of the RGBD input, we construct MPI at \emph{scene-specific} depth by designing a novel \emph{Plane Adjustment Network} to adjust the depth of the planes from an initial configuration using an attention mechanism. 
Compared to a scene-agnostic MPI with predefined depth, our method can better fit the geometry and appearance of the scene, leading to superior view synthesis quality with fewer visual distortions. 
A similar concept of variable depth was proposed in VMPI~\cite{Li2020LF}, where the output for each plane contains RGB$\alpha$ and an extra depth channel. However, this solution suffers from even more severe over-parameterization and difficulty in MPI learning. Our depth adjustment scheme is significantly different from VMPI with the goal of overcoming the issues induced by over-parameterization.

With adjusted plane depths, a \emph{Color Prediction Network} is applied to generate the color and density values for the planes.
Our Color Prediction Network uses an encoder-decoder architecture similar to \cite{mine2021}. 
Specifically, the encoder encodes the RGBD image to features shared by each plane. The decoder predicts the color for a \emph{single} plane given its depth and the shared features. Such a color generation scheme avoids generating a large number of output channels at once thus is easier to train and generalize to test data. However, if handled naively, the color prediction for each plane is independent and agnostic to other planes, which is suboptimal for our case with dynamically-adjusted depth positions. 
Therefore, we further propose a novel \emph{feature masking mechanism} to introduce proper inter-plane interactions in color prediction, which is crucial to our adaptive MPI.

To train our method, we also present a \emph{warp-back} strategy to generate stereo training pairs from unconstrained image datasets such as COCO~\cite{coco2018cvpr}. No stereo or multi-view images captured in real life or rendered with graphics engine were used for training.
Our method is tested on multiple datasets including both synthetic and real ones.
The experiments show that it can achieve superior view synthesis quality and generalization ability, outperforming the previous methods by a wide margin.

In summary, our contributions include:
\begin{itemize}
    \item We propose a \emph{AdaMPI} architecture to address the single-view view synthesis problem in the wild. It contains a novel Plane Adjustment Network to predict adaptive and scene-specific MPI depth and a Color Prediction Network with a novel feature-masking scheme to predict the color of each plane in a separate but interactive manner. 
    \item We construct large-scale stereo training data using only unconstrained single-view image collections by a simple yet effective \emph{warp-back} strategy.
    \item We demonstrate state-of-the-art results on various datasets with a single trained model.
\end{itemize}

\section{Related Work}
\subsection{Neural Scene Representations}
Emerging works integrate scene representation into a neural network and optimize it using 2D multi-view supervision~\cite{aliev2020neural,lombardi2019neural,Riegler2021SVS,liu2020neural,dai2020neural,mildenhall2020nerf,Kellnhofer:2021:nlr,tulsiani2018layer}.
At test time, they can use it to render novel views with geometric consistency.
Implicit representations, such as the neural radiance field~\cite{mildenhall2020nerf}, offer the potential to model complex geometry and reflectance by representing a scene as a continuous function of color and density, but they are difficult to generalize to arbitrary unseen scenes.
Although some recent works propose to improve its generalization ability~\cite{yu2021pixelnerf,wang2021ibrnet}, the performance is still far from satisfying, especially for the single-view setup.
Explicit representations are also used for view synthesis, such as point cloud~\cite{wiles2020synsin,Rockwell2021}, mesh~\cite{hu2021worldsheet}, and voxel~\cite{lai2021video,nguyen2019hologan}.
In this paper, we adopt multiplane images (MPI)~\cite{zhou2018stereo} as our 3D scene representation, which not only has strong representation capability as demonstrated by previous methods but also enjoys fast rendering speed.

\subsection{Multiplane Images}
A standard MPI consists of $N$ fronto-parallel RGB$\alpha$ planes predefined in the camera's view frustum \cite{zhou2018stereo}.
It is first used for view synthesis from two or more views \cite{zhou2018stereo,srinivasan2019pushingmpi,flynn2019deepview} and later applied to handle single-view input~\cite{single_view_mpi} for certain scene categories.
MPI is known to be a highly over-parameterized representation \cite{zhou2018stereo}.  A vanilla MPI is a $H\times W\times 4N$ tensor, where $H$ and $W$ are the image height and width and $N$ is the plane number. 
Previous works proposed to reduce the number of output channels by reusing the RGB channel~\cite{zhou2018stereo,single_view_mpi}, which inevitably sacrifices the representation capability.
A recent work of \cite{mine2021} proposes to apply an encoder to extract shared features a single forward pass and a decoder that runs $N$-times to predict the RGB$\alpha$ channels for each plane.
Although better results are achieved, it predicts each plane independently, which is sub-optimal for the MPI representation especially when the plane positions can vary across scenes. In this regard, most previous works use a fixed set of MPI planes at predefined depth for simplification~\cite{zhou2018stereo,single_view_mpi,mine2021,srinivasan2019pushingmpi}, which is clearly not ideal to handle diverse scene layouts in the wild. An exception is VMPI~\cite{Li2020LF}, where the output for each plane contains RGB$\alpha$ and an extra depth channel. However, the naively-added depth channel leads to heightened over-parameterization and poorer performance when plane number increases. 
A recent work~\cite{luvizon2021adaptive} concurrent to ours proposes to construct MPI planes at heuristically-selected depth, which is significantly different from our learnable depth adjustment scheme.

\subsection{Single-View View Synthesis}
Synthesizing novel views from a single image is a highly ill-posed problem.
One popular scheme to handle this problem is to train a network that predicts a 3D scene representation from a single input image and optimize it using multi-view supervision~\cite{wiles2020synsin,single_view_mpi,lai2021video,hu2021worldsheet,mine2021,Rockwell2021,srinivasan2017learning,rombach2021geometry}.
An additional network is often learned to estimate depth maps or apply loss terms to depth.
However, these methods are difficult to generalize to in-the-wild scenes due to the lack of large-scale multi-view datasets.
In this paper, we propose a novel strategy to generate training data from single-view image collections.
Another line of works leverages single-view depth estimation to decompose the scene into multiple layers and learns an inpainting network~\cite{Nazeri_2019_ICCV,yu2019free} to extend each occluded layer \cite{Shih3DP20,jampani:ICCV:2021}.
These methods do not need multi-view training data and can handle in-the-wild scenes.
However, the layered depth representation used by these methods are sensitive to depth discontinues which determine  depth layering and they have difficulty addressing complex 3D scene structures.

\section{Method}
Given an input image $I_s$ and the corresponding depth map $D_s$ obtained from a monocular depth estimation system, our method generates an explicit multiplane images (MPI) representation from which novel views can be efficiently rendered.

\subsection{Preliminaries: Multiplane Images} \label{Sec:Method:MPI}
An MPI consists of $N$ fronto-parallel RGB$\alpha$ planes in the frustum of the source camera with viewpoint $v_s$, arranged at depths $d_1,...,d_N$ for planes from the nearest to farthest.
Most previous works~\cite{zhou2018stereo,mine2021} use a predefined set of $\{d_i\}$, whereas our method predicts $\{d_i\}$ by the \emph{Plane Adjustment Network}, to be described later.
Let the color and alpha channel of $i$-th plane be $c_i$ and $\alpha_i$ respectively, then each plane can be represented as $(c_i,\alpha_i,d_i)$.

Given a target viewpoint $v_t$, an image can be rendered from the source-view MPI in a differentiable manner using planar inverse warping. Specifically, each pixel $[u_t,v_t]$ on the target-view image plane can be mapped to pixel $[u_s,v_s]$ on $i$-th source-view MPI plane via homography function~\cite{Hartley2004}:
\begin{equation}
    [u_s, v_s, 1]^T \sim K_s(R-\frac{tn^T}{d_i})K_t^{-1}[u_t, v_t, 1]^T,
    % \vspace{-2pt}
\end{equation}
where $R$ and $t$ are the rotation and translation, $K_s$ and $K_t$ are the camera intrinsics, and $n=[0,0,1]^T$ is the normal vector of the planes in the source view. Then, the color and alpha $c_i'$ and $\alpha_i'$ can be obtained via bilinear sampling and composited using the over operation~\cite{alpha_comp} to render the image $\widehat{I_t}$ at $v_t$:
\begin{equation}
    \widehat{I_t}={\sum}_{i=1}^N \left( c_i' \alpha_i' {\prod}_{j=1}^{i-1}(1-\alpha_j') \right).
    \label{eq:alphacomp}
\end{equation}

In our implementation, we predict density  $\sigma_i$ for each plane instead of alpha $\alpha_i$.
We convert $\sigma_i$ to $\alpha_i$ according to the principles from classic volume rendering as in \cite{mine2021}:
\begin{equation}
    \alpha_i=\exp\left( -\delta_i\sigma_i \right),
    \label{eq:sigma2alpha}
\end{equation}
where $\delta_i$ is the distance map between plane $i$ and $i+1$ as in \cite{mine2021}.
We empirically find this parameterization leads to sharper results in our method.

\begin{figure*}[t]
    \centering
    \includegraphics[width=1\textwidth]{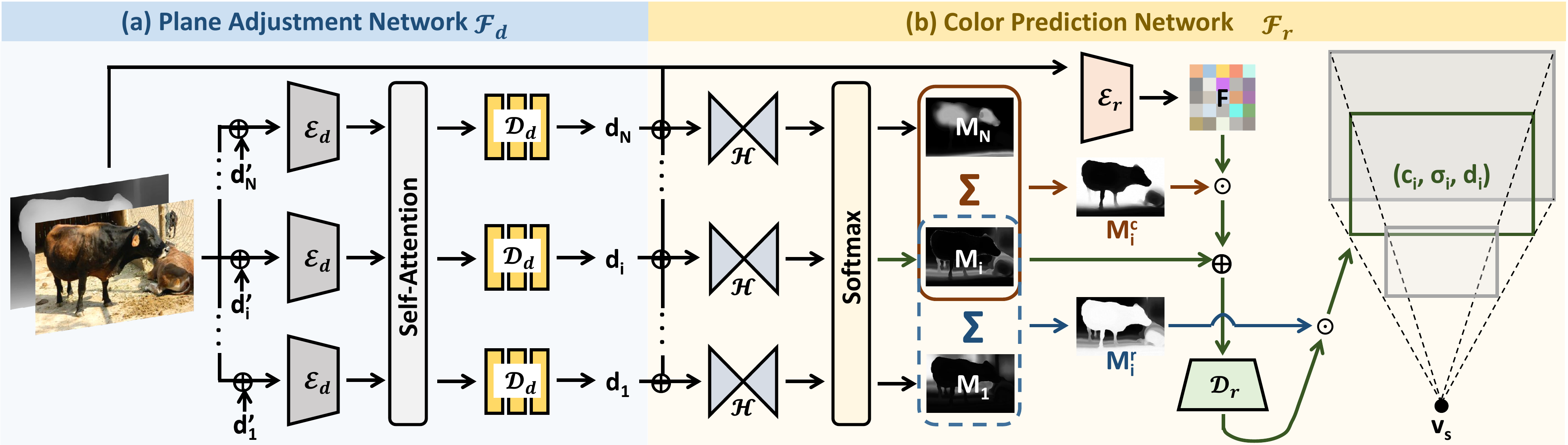}
    \vspace{-17pt}
    \caption{Overview of our framework, which consists of two components: (a) the Plane Adjustment Network $\mathcal{F}_d$ adjusts an initial predefined MPI depth $\{d_i'\}_{i=1}^N$ to a scene-specific one $\{d_i\}_{i=1}^N$ according to the geometry and appearance features; (b) the Color Prediction Network $\mathcal{F}_r$ predicts the color $c_i$ and density $\sigma_i$ for each plane at $d_i$.
    Here, $\oplus$ and $\odot$ denote  concatenation and element-wise multiplication; each scalar depth is repeated $H\times W$ times before concatenation.}
    \label{Fig:framework}
\end{figure*}

\subsection{Network Architecture} \label{Sec:Method:net}
The goal of our network $\mathcal{F}$ is to predict $N$ planes each with color channels $c_i$, density channel $\sigma_i$, and depth $d_i$ from an input image $I_s$ and its depth map $D_s$:
\begin{equation}
    \{(c_i,\sigma_i,d_i)\}_{i=1}^N=\mathcal{F}(I_s,D_s).
\end{equation}
The depth maps are obtained from an off-the-shelf monocular depth estimation network~\cite{Ranftl2021}.
As illustrated in Figure~\ref{fig:teaser},  $\mathcal{F}$ has two sub-networks: a \emph{Plane Adjustment Network} $\mathcal{F}_d$ and a \emph{Color Prediction Network} $\mathcal{F}_r$.
We apply $\mathcal{F}_d$ to infer the plane depth $\{d_i\}_{i=1}^N$ and $\mathcal{F}_r$ to predict the color and density at each $d_i$:
\begin{equation}
    \{d_i\}_{i=1}^N = \mathcal{F}_d(I_s,D_s), \;\; \{(c_i,\sigma_i)\}_{i=1}^N = \mathcal{F}_r(I_s,D_s,\{d_i\}_{i=1}^N).
\end{equation}

\subsubsection{Plane Adjustment Network} \label{Sec:Method:net:DPN}
Given  $I_s$ and $D_s$ at the source view, our \emph{Plane Adjustment Network (PAN)}  $\mathcal{F}_d$ is responsible for arranging each MPI plane at an appropriate depth to represent the scene.
A straightforward way is to apply an RGBD encoder to directly predict the MPI depth.
However, it can only handle a fixed number of MPI planes once trained.
Therefore, we propose to adjust the MPI depth from an initial predefined sampling $\{d_i'\}_{i=1}^N$. 
As illustrated in Fig.~\ref{Fig:framework}(a), we first use a shared lightweight CNN $\mathcal{E}_d$ to extract a global feature $f_i'$ for plane $i$ at the initial depth $d_i'$:
\begin{equation}
    f_i' = \mathcal{E}_d(I_s, D_s, d_i').
\end{equation}
Then we apply the self-attention operation~\cite{vaswani2017attention} to $\{f_i'\}_{i=1}^N$ to obtain $\{f_i\}_{i=1}^N$:
\begin{equation}
    \{f_i\}_{i=1}^N =  \textit{Self-Attention}\big(\{f_i'\}_{i=1}^N\big)
\end{equation}
The intuition here is to adjust the MPI depth at the feature level by exchanging the geometry and appearance information among $\{f_i'\}_{i=1}^N$.
The adjusted feature $f_i$ is then decoded to the adjusted depth $d_i$ using a shared MLP $\mathcal{D}_d$:
\begin{equation}
    d_i = \mathcal{D}_d(f_i).
\end{equation}
In our implementation, we set the initial depth of MPI planes uniformly spaced in disparity as in \cite{single_view_mpi}.

\subsubsection{Radiance Prediction Network} \label{Sec:Method:net:RN}
Given adjusted plane depths, our \emph{Color Prediction Network (CPN)} $\mathcal{F}_r$ produces the color and density channels for each plane. To achieve this, $\mathcal{F}_r$ should properly interpret the scene structure to represent the visible pixels on the planes and inpaint the occluded content.
We use an RGBD encoder to encode the scene structure and propose a novel \emph{feature mask mechanism} to facilitate the decoder to predict color and density attributes for both the visible and the occluded pixels.

As illustrated in Fig.~\ref{Fig:framework}(b), we first apply an encoder $\mathcal{E}_r$ to encode the source view to a set of feature maps $F$ shared across planes:
\begin{equation}
    F = \mathcal{E}_r(I_s,D_s).
\end{equation}
%Then, the \emph{feature mask mechanism} guides the decoder $\mathcal{D}_r$ to produce the color and density channel for each plane from $F$.
To guide the decoder $\mathcal{D}_r$ in color and density prediction for each plane, we introduce three types of masks described below.

\begin{figure*}[t]
    \centering
    \includegraphics[width=1\textwidth]{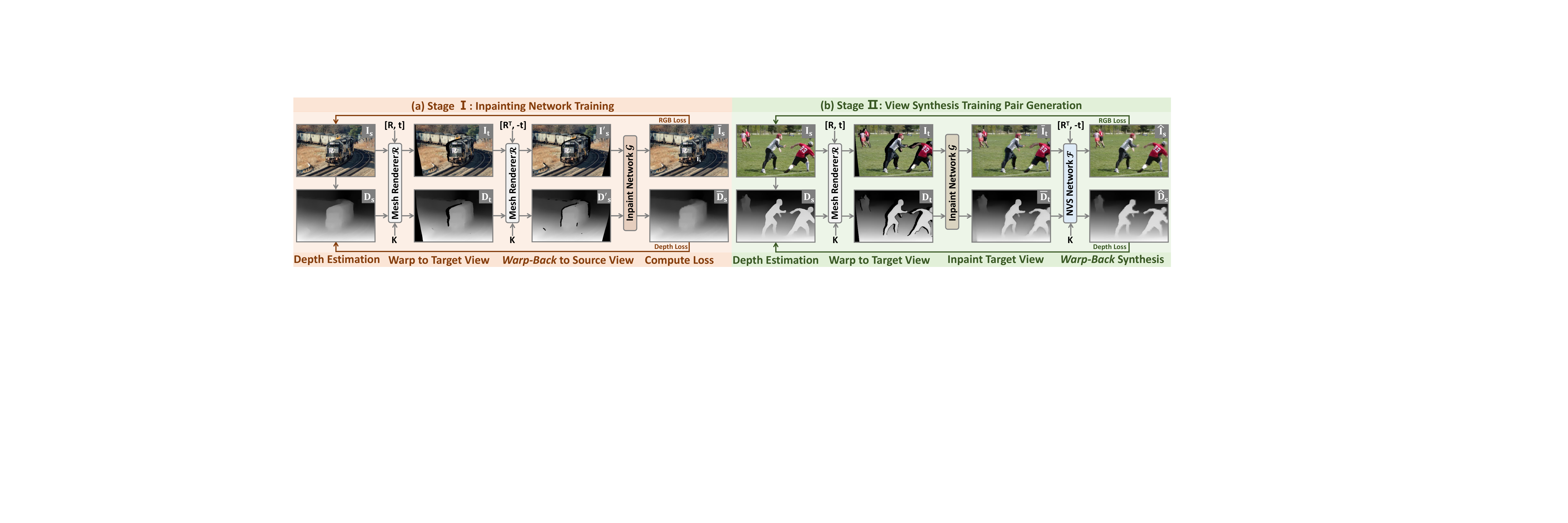}
    \vspace{-19pt}
    \caption{Overview of our \emph{warp-back} strategy to train novel view synthesis method using only single-view images.
    (a) We first train a network $\mathcal{G}$ specialized to inpaint the holes caused by view change. The holes are generated by \emph{back-warping}. (b) We generate stereo image pairs from a single image to train the novel view synthesis network $\mathcal{F}$. The warped and inpainted images are used as input and the original images are the target, which is also a \emph{back-warping} setup. }
    \label{Fig:warpback}
\end{figure*}

The \textbf{\emph{Feature Mask}} $M_i$ assigns each visible pixel in the source view softly to each plane. 
We employ a UNet-like network $\mathcal{H}$ followed by a softmax layer to generate a feature mask for each plane:
\begin{equation}\label{eq:softmax}
    \{{M_i}\}_{i=1}^N = \emph{Softmax}(\{\mathcal{H}(I_s,D_s,d_i)\}_{i=1}^N).
\end{equation}

The \textbf{\emph{Context Mask}} $M_i^c$ represents the context regions for each plane that can be used to inpaint the occluded pixels. 
Since the occluded pixels are irrelevant to the occluding content in the front, we define the context mask $M_i^c$ as the union of the pixels on and behind the $i$-th plane:
\begin{equation}
    M_i^c = {\sum}_{j=i}^N M_j.
\end{equation}

The \textbf{\emph{Rendering Mask}} $M_i^r$ is used to clean the unwanted information in regions behind the $i$-th plane while retaining the visible and inpainted pixels:
% \vspace{-1pt}
\begin{equation}
    M_i^r = {\sum}_{j=1}^i M_j.
\end{equation}

As shown in Fig.~\ref{Fig:framework}(b), we first multiply the shared feature maps $F$ with the context mask $M_i^c$ to retrieve the context information for the $i$-th plane for better occlusion inpainting.
Then we concatenate it with the feature mask $M_i$ to softly assign the visible pixels to the plane.
Next, we send it to the decoder $\mathcal{D}_r$ and multiply the predicted channels with the rendering mask $M_i^r$ to clean up the background information and obtain the final color and density. 
The above process can be written as:
\begin{gather}
    (c_i,\sigma_i) = M_i^r \odot \mathcal{D}_r(M_i^c \odot F \;\; || \;\; M_i),
\end{gather}
where $||$ denotes the concatenation operation.

In our method, the encoder $\mathcal{E}_r$ only run once while the feature mask network $\mathcal{H}$ and the decoder $\mathcal{D}_r$ run $N$ times for the $N$ planes.
This architecture design can effectively mitigate the over parameterization problem of MPI by reducing the output channel numbers of the network. Unlike \cite{mine2021} which predicts each plane independently, we introduce inter-plane interactions by injecting the masks, which is important in our case where the plane positions are varying.
We implement $\mathcal{E}_r$ with a ResNet-18 structure~\cite{he2016deep}, $\mathcal{H}$ a UNet-like architecture~\cite{ronneberger2015u}, and $\mathcal{D}_r$ a Monodepth2-like structure~\cite{godard2019digging} with gated convolution~\cite{yu2019free}.

\subsection{Training Data} \label{Sec:Method:Data}
To train our network, multiview images are needed as in previous view synthesis works~\cite{single_view_mpi,wiles2020synsin,mine2021}.
To handle in-the-wild photos, the training set should contain a wide range of scene types.
However, creating such a large-scale multiview dataset is prohibitively laborious and no existing one meets the above criterion to our knowledge.
On the other hand, large-scale single image datasets~\cite{coco2018cvpr} are much easier to collect.
This motivates us to leverage the single-view images to generate training pairs.
Inspired by~\cite{watson-2020-stereo-from-mono,Aleotti_CVPR_2021}, we warp the source image to a random target view according to the estimated depth to synthesize multiview images. A \emph{warp-back} strategy is applied, which is the key to generate the data to train our inpainting and view synthesis networks.

\begin{figure}[t]
\vspace{2pt}
    \centering
    \includegraphics[width=1.0\columnwidth]{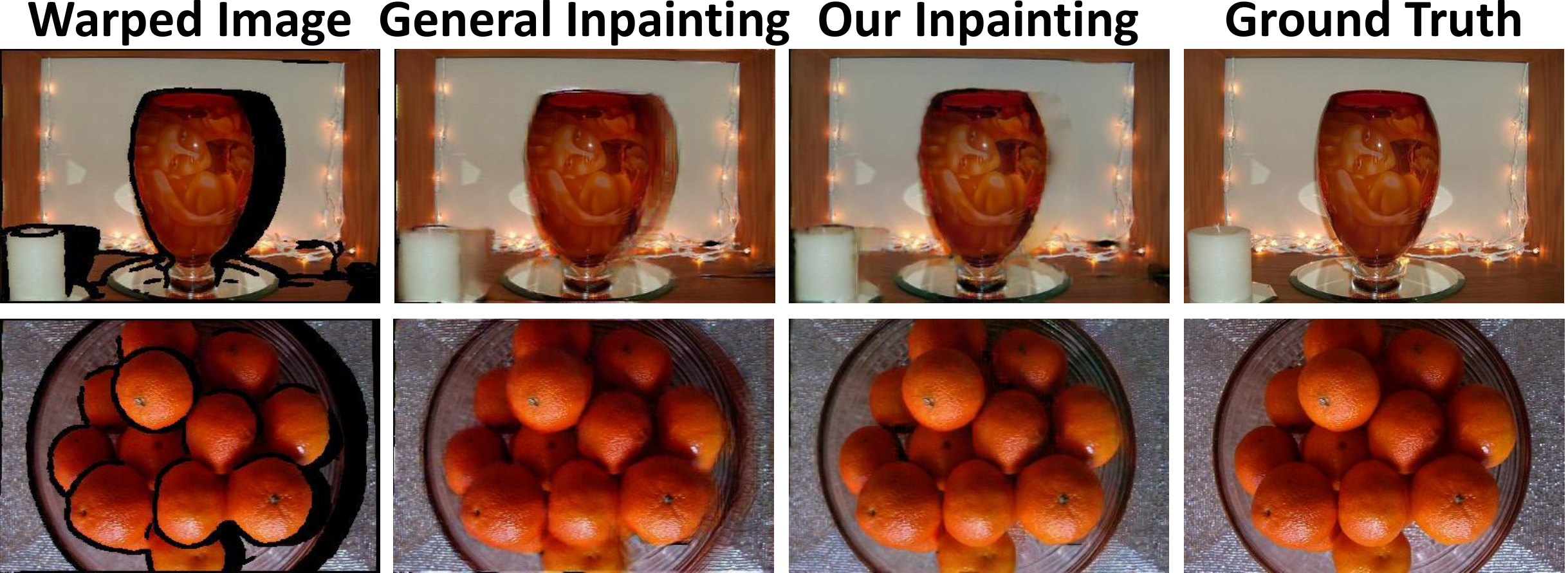}
    \vspace{-18pt}
    \caption{Comparison of a general-purpose inpainting network trained on random masks and our network specialized for holes caused by view change.}
    \label{Fig:inpaint_compare}
\end{figure}

\subsubsection{Depth-based Warping} \label{Sec:Method:Data:target-view-rendering}
First, we use a mesh renderer $\mathcal{R}$ to render a random target view given the source image $I_s$ and its depth map $D_s$ from monocular depth estimation.
Specifically, we generate a plausible intrinsic $K$ and camera motion $(R,t)$ as in \cite{Aleotti_CVPR_2021}, where $K$ is shared by source and target view and $(R, t)$ describes the transformation from the source view to the target view.
A mesh is created by lifting the pixel to 3D vertices based on to their depth and connecting them as a triangle mesh according to their affinity on image grid. 
%Similar to~\cite{infinite_nature_2020}, we removing long  edges by shareholding.
We remove long edges by thresholding the gradients on the depth map. The mesh is then rendered to the target view to obtain image $I_t$ and depth $D_t$.

\subsubsection{Back-warping and Inpainting Network Training} \label{Sec:Method:Data:int}
The holes in the rendered $I_t$ and $D_t$ represent the unseen content in the source view.
Since the distribution of these holes induced by camera view change is quite different from the random masks used by the general inpainting  network~\cite{Nazeri_2019_ICCV}, we opt to train a network that is specialized to inpaint these holes. 
In this way, we can not only improve the inpainting quality through targeted training, but also reduce the domain gap between the generated training data and the real test cases. 

To achieve this, we use a \emph{warp-back} strategy to generate training pairs for the inpainting network. As illustrated in Fig.~\ref{Fig:warpback}(a), we warp $I_t$ and $D_t$ \emph{back to the source view} to obtain $I_s'$ and $D_s'$.
Then we train a network $\mathcal{G}$ to inpaint the holes in $I_s'$ and $D_s'$ with $I_s$ and $D_s$ serving as the ground truth.
To enhance the alignment between the inpainted depth map $\bar{D_s}$ and color image $\bar{I_s}$, we adopt the EdgeConnect~\cite{Nazeri_2019_ICCV} architecture as in~\cite{Shih3DP20}. 
Specifically, we first apply a network to inpaint edges in the holes and then adopt two separate networks to inpaint color and depth based on the inpainted edges.
The inpainting networks are trained 
using the default settings in \cite{Nazeri_2019_ICCV}.
Figure~\ref{Fig:inpaint_compare} shows that our network successfully borrows context information from the background to inpaint the holes, whereas a general-purpose  inpainter \cite{Nazeri_2019_ICCV} fails to handle these cases. 

\subsubsection{View Synthesis Training Pair Generation} \label{Sec:Method:Data:tpg}
During the training process for our view synthesis network, we generate stereo image pairs on the fly.
As illustrated in Fig.~\ref{Fig:warpback}(b), we first sample an image $I_s$ with depth map $D_s$ and generate a plausible intrinsic $K$ and camera motion $(R, t)$.
We then render the target color image $I_t$ and depth map $D_t$ and apply the our trained inpainter $\mathcal{G}$ to fill the holes, which gives rise to $\bar{I_t}$ and $\bar{D_t}$.
Optionally, one can also pre-generate a large-scale training set offline.
To ensure the network receives supervision from real image distribution, we adopt $(\bar{I}_t,\bar{D}_t)$ as the input to $\mathcal{F}$ and $(I_s,D_s)$ as the ground truth target view, as shown in Fig.~\ref{Fig:warpback}(b), which follows a similar \emph{back-warping} spirit. 

\subsection{Network Training} \label{Sec:Method:Train}
The training pairs for our method contain the color images and depth maps of the source and target views and the camera parameters: $(I_s,I_t,D_s,D_t,K,R,t)$.
Our overall loss function to train the network combines a view synthesis term $\mathcal{L}_{vs}$ and a regularization term $\mathcal{L}_{reg}$.
The goal of $\mathcal{L}_{vs}$ is to encourage the rendered target view color image and depth map to match the ground truth.
We employ L1 loss, SSIM loss, perceptual loss~\cite{chen2017photographic}, and focal frequency loss~\cite{jiang2021focal} on the rendered $\widehat{I_t}$ and L1 loss on the rendered $\widehat{D_t}$ with weight $1,1,0.1,10$, respectively.
In the regularization term $\mathcal{L}_{reg}$, we introduce a rank loss to regularize the MPI depth predicted by $\mathcal{F}_d$ to be in a correct order, and an assignment loss to enforce $\mathcal{F}_d$ and $\mathcal{H}$ to produce reasonable results:

\begin{equation}
    \mathcal{L}_{reg} = \lambda_{rank} \mathcal{L}_{rank} + \lambda_{assign} \mathcal{L}_{assign}, {\rm where}
\end{equation}

\begin{equation}
    \mathcal{L}_{rank} = \frac{1}{N-1}{\sum}_{i=1}^{N-1} \max(0, d_{i+1}-d_{i}),
\end{equation}
    
\begin{equation}
    \mathcal{L}_{assign} = \frac{1}{HW}{\sum}_{i=1}^N {\sum}_{(x,y)} M_i\odot|D_s-d_i|.
\end{equation}

Intuitively, $\mathcal{L}_{assign}$ measures the error to represent the depth map $D_s$ using $N$ discrete planes at $\{d_i\}_{i=1}^N$ with masks $\{M_i\}_{i=1}^N$ (note that $\sum_i M_i = 1$; see Eq.~\ref{eq:softmax}).
The loss weights are set as $\mathcal{L}_{rank}=100$ and $\mathcal{L}_{assign}=10$. We use the Adam optimizer~\cite{kingma2014adam}  with learning rate $0.0001$ and batch size $12$ for training.

\section{Experiments} \label{Sec:Exp}
\subsection{Experimental Setup}
\subsubsection{Datasets} 
To generate our stereo training data, we use the training set of COCO~\cite{coco2018cvpr} which contains 111K still images. The state-of-the-art monocular depth estimator DPT~\cite{Ranftl2021} is applied to obtain the corresponding depth maps. 
For quantitative evaluation, 
we use four datasets that provide multi-view images or videos of a static scene: Ken Burns~\cite{Niklaus_TOG_2019}, TartanAir~\cite{wang2020tartanair}, RealEstate-10K~\cite{zhou2018stereo}, and Tank \& Temples~\cite{Knapitsch2017}. 
See suppl. document for details on data selection and data processing. 

\subsubsection{Baselines}
We compare our method with 3D-Photo~\cite{Shih3DP20}, SLIDE~\cite{jampani:ICCV:2021}, and several MPI-based methods: SVMPI~\cite{single_view_mpi}, MINE~\cite{mine2021}, and VMPI~\cite{Li2020LF}.
For 3D-Photo, we use the trained model provided by the authors for evaluation.
For SVMPI and MINE, since their original methods take only a color image as input, we train an RGBD version from scratch using our dataset and loss functions. 
We denote the modified and retrained models as SVMPI++ and MINE++.
For VMPI, we directly take their network  and retrain it using our dataset. 
We use the same depth maps and camera parameters for all methods to ensure fair comparison.
For SLIDE, we qualitatively compare our method with it in Fig.~\ref{Fig:visual_slide} using an example from their paper as their code is not available.

\subsubsection{Metrics}
We report the SSIM, PSNR, and LPIPS~\cite{zhang2018perceptual} scores to measure the rendering quality.  
We crop $5\%$ pixels on the border before evaluation following \cite{single_view_mpi,mine2021} since we do not handle large out-of-fov inpainting.

\begin{figure}
        \centering
        \includegraphics[width=1.0\columnwidth]{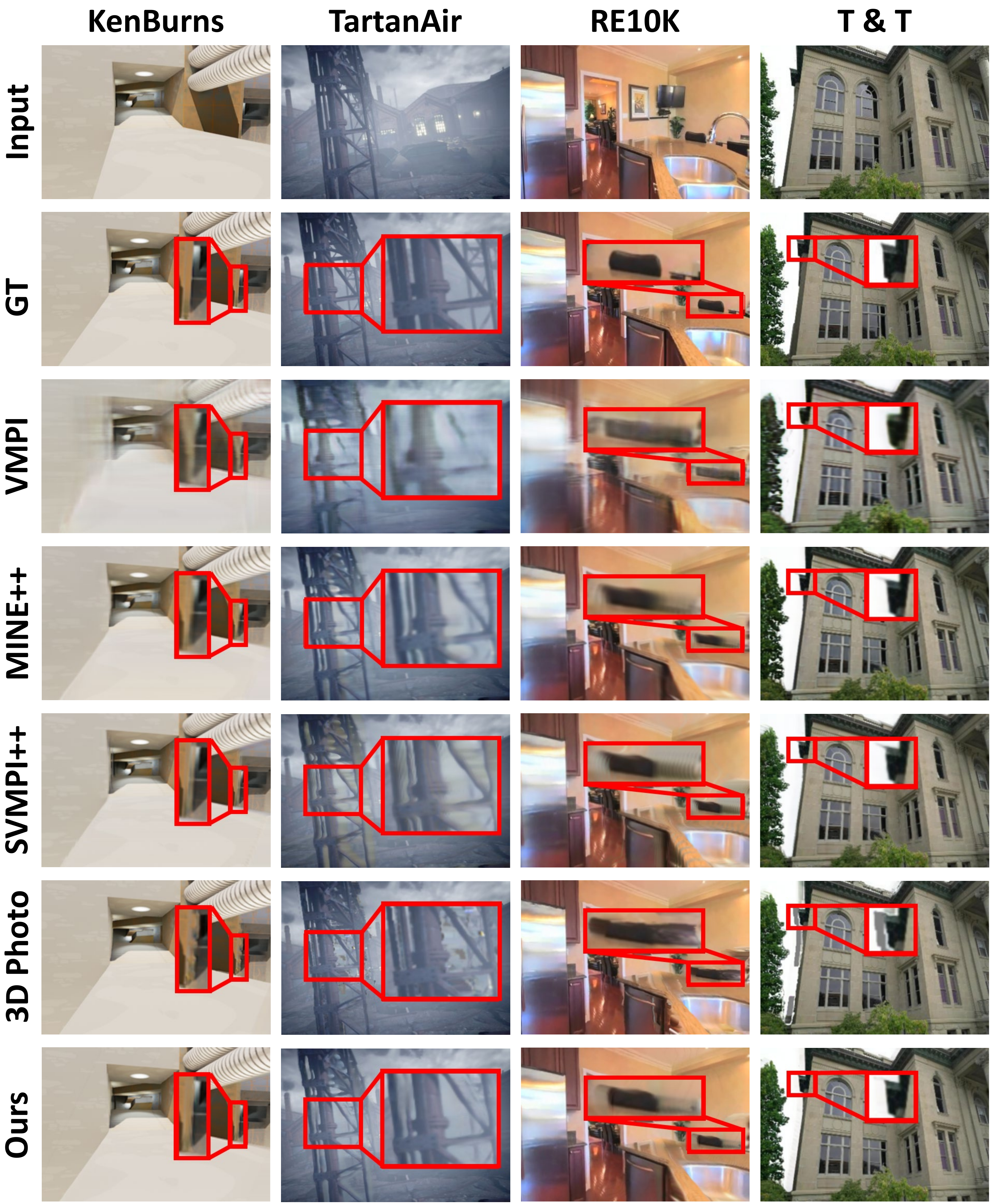}
        \vspace{-19pt}
        \caption{Qualitative comparison between our novel view synthesis method and other approaches on different datasets.
        (\textbf{Best viewed with zoom-in.})}
        \label{Fig:visual_benchmark}
        \vspace{-2pt}
    \end{figure}

\begin{figure}[t]
    \centering
    \includegraphics[width=1.0\columnwidth]{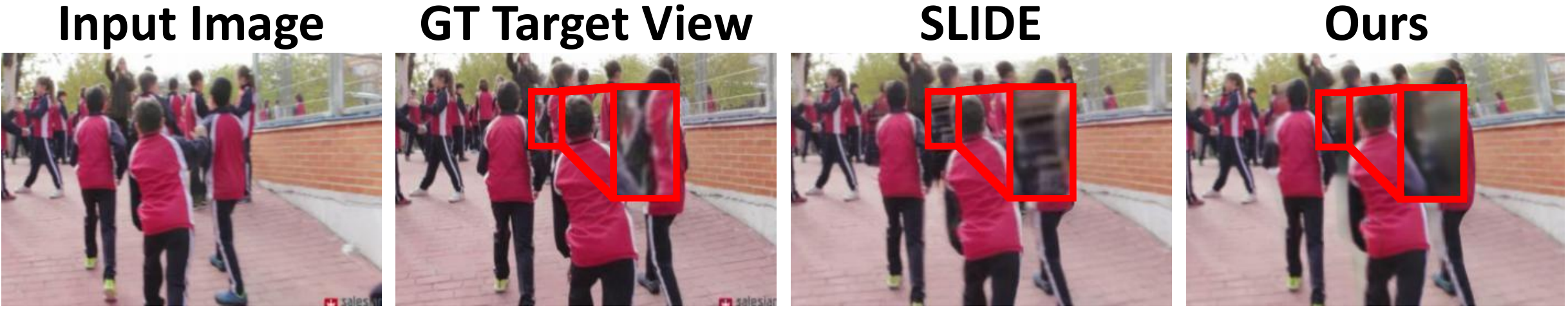}
    \vspace{-19pt}
    \caption{Novel view synthesis results from our method and SLIDE using the image sample from SLIDE. 
    For a scene with complex occlusions, the SLIDE method that uses two depth layers generates texture-stretching artifact. Our method inpaints the occluded region with contents consistent with the background texture. (\textbf{Best viewed with zoom-in.})
    }
    \label{Fig:visual_slide}
    \vspace{-3pt}
\end{figure}

\subsection{Comparison with Previous Methods}
The performance of MPI-based methods is related to the number of planes $N$. Intuitively, better results should be obtained with more planes; but \emph{this is not the case for all methods}. We first identify the best $N$ for each MPI-based method using the Ken Burns dataset where ground-truth depths are available.  Table~\ref{tab:vary_plane} presents the results of different methods with 8, 16, 32, and 64 planes, respectively.
As we can see, our method achieves consistently better quantitative results
compared to other MPI-based methods with varying $N$. The performance of VMPI decreases significantly when using $N=16$ planes, as their representation is highly over-parameterized ($5N$ output channels) and a CNN is applied to directly generate such a large tensor. 
SVMPI++ reduces the number of output channels by predicting only one color image, but 
the channel number is still in proportion to $N$.
Their performance decreases when increasing the plane number to $N=64$.
MINE++ and our method enjoy stable performance gain with increasing numbers of planes.

We then benchmark 3D-Photo and all the MPI-based methods on all four datasets.
We test each MPI-based method \emph{with the best plane number $N$ according to  Table~\ref{tab:vary_plane}}, 
\emph{i.e.}, 8 planes for VMPI, 32  for SVMPI++, and 64 for MINE++ and our method.
Table~\ref{table:benchmark} shows that our method significantly outperforms the others, especially on the former three datasets.
Figure~\ref{Fig:visual_benchmark} shows some qualitative results.
Compared to VMPI, our AdaMPI is easier to train and produces significantly better image quality.
Compared to SVMPI++ and MINE++, our method generates sharper and more realistic results since our scene-specific MPI can better represent thin structures and sloped surfaces.
Besides, our feature masking scheme explicitly assigns visible pixels to each plane, which reduces the common repeated-texture artifacts~\cite{srinivasan2019pushingmpi}. 
Compared to 3D-Photo, our results 
have fewer artifacts around depth discontinues.

\begin{table}[t]
\centering
\scriptsize
\caption{Quantitative comparison between our method and other MPI-based approaches with varying number of planes on the Ken Burns dataset.}
\vspace{-8pt}
\begin{tabular}{lcccccccccccc}
\toprule
& \multicolumn{3}{c}{$N=8$}
& \multicolumn{3}{c}{$N=16$} 
& \multicolumn{3}{c}{$N=32$} 
& \multicolumn{3}{c}{$N=64$}
\\
\cmidrule(lr){2-4}
\cmidrule(lr){5-7}
\cmidrule(lr){8-10}
\cmidrule(lr){11-13}
\!\!\!Method\!\!\!
& \!\!\!\!\!LPIPS\ \!\!$\downarrow$\!\!\!\!\! & \!\!\!\!PSNR\ \!\!$\uparrow$\!\!\!\! & \!\!\!\!SSIM\ \!\!$\uparrow$\!\!\!\!
& \!\!\!\!\!LPIPS\ \!\!$\downarrow$\!\!\!\!\! & \!\!\!\!PSNR\ \!\!$\uparrow$\!\!\!\! & \!\!\!\!SSIM\ \!\!$\uparrow$\!\!\!\!
& \!\!\!\!\!LPIPS\ \!\!$\downarrow$\!\!\!\!\! & \!\!\!\!PSNR\ \!\!$\uparrow$\!\!\!\! & \!\!\!\!SSIM\ \!\!$\uparrow$\!\!\!\!
& \!\!\!\!\!LPIPS\ \!\!$\downarrow$\!\!\!\!\! & \!\!\!\!PSNR\ \!\!$\uparrow$\!\!\!\! & \!\!\!\!SSIM\ \!\!$\uparrow$\!\!\!\!
\\ 
\midrule
\!\!\!VMPI\!\!\! & \!\!\!0.333\!\!\! & \!\!\!26.57\!\!\! & \!\!\!0.820\!\!\! &  \!\!\!0.387 \!\!\! &  \!\!\!24.12 \!\!\! &  \!\!\!0.677 \!\!\! & \!\!\!0.362\!\!\! & \!\!\!24.68\!\!\! & \!\!\!0.767\!\!\! & \!\!\!0.369\!\!\! & \!\!\!24.22\!\!\! & \!\!\!0.757\!\!\! \\
\!\!\!SVMPI++\!\!\! & \!\!\!0.126\!\!\! & \!\!\!29.09\!\!\! & \!\!\!0.878\!\!\! & \!\!\!0.092\!\!\! & \!\!\!31.86\!\!\! & \!\!\!0.932\!\!\! & \!\!\!0.094\!\!\! & \!\!\!32.32\!\!\! & \!\!\!0.946\!\!\! & \!\!\!0.099\!\!\! & \!\!\!31.90\!\!\! & \!\!\!0.939\!\!\! \\
\!\!\!MINE++\!\!\! & \!\!\!0.132\!\!\! & \!\!\!29.12\!\!\! & \!\!\!0.877\!\!\! & \!\!\!0.120 \!\!\! &  \!\!\!31.20 \!\!\! &  \!\!\!0.925 \!\!\! & \!\!\!0.128\!\!\! & \!\!\!31.52\!\!\! & \!\!\!0.938\!\!\! & \!\!\!0.117\!\!\! & \!\!\!31.40\!\!\! & \!\!\!0.940\!\!\! \\
\!\!\!\emph{Ours}\!\!\! & \!\!\!\textbf{0.099}\!\!\! & \!\!\!\textbf{30.31}\!\!\! & \!\!\!\textbf{0.902}\!\!\! & \!\!\!\textbf{0.073}\!\!\! & \!\!\!\textbf{32.76}\!\!\! & \!\!\!\textbf{0.942}\!\!\! & \!\!\!\textbf{0.061}\!\!\! & \!\!\!\textbf{34.19}\!\!\! & \!\!\!\textbf{0.962}\!\!\! & \!\!\!\textbf{0.059}\!\!\! & \!\!\!\textbf{34.21}\!\!\! & \!\!\!\textbf{0.966}\!\!\! \\
\bottomrule
\end{tabular}
\label{tab:vary_plane}
\vspace{-2pt}
\end{table}

\begin{table}[t]
\centering
\scriptsize
\caption{Quantitative comparison between our method and previous approaches on four multi-view datasets.}
\vspace{-8pt}
\begin{tabular}{lcccccccccccc}
\toprule
& \multicolumn{3}{c}{Ken Burns}
& \multicolumn{3}{c}{TartanAir} 
& \multicolumn{3}{c}{RealEstate-10K} 
& \multicolumn{3}{c}{Tank \& Temples}
\\
\cmidrule(lr){2-4}
\cmidrule(lr){5-7}
\cmidrule(lr){8-10}
\cmidrule(lr){11-13}
\!\!\!Method\!\!\!
& \!\!\!\!\!LPIPS\ \!\!$\downarrow$\!\!\!\!\! & \!\!\!\!PSNR\ \!\!$\uparrow$\!\!\!\! & \!\!\!\!SSIM\ \!\!$\uparrow$\!\!\!\!
& \!\!\!\!\!LPIPS\ \!\!$\downarrow$\!\!\!\!\! & \!\!\!\!PSNR\ \!\!$\uparrow$\!\!\!\! & \!\!\!\!SSIM\ \!\!$\uparrow$\!\!\!\!
& \!\!\!\!\!LPIPS\ \!\!$\downarrow$\!\!\!\!\! & \!\!\!\!PSNR\ \!\!$\uparrow$\!\!\!\! & \!\!\!\!SSIM\ \!\!$\uparrow$\!\!\!\!
& \!\!\!\!\!LPIPS\ \!\!$\downarrow$\!\!\!\!\! & \!\!\!\!PSNR\ \!\!$\uparrow$\!\!\!\! & \!\!\!\!SSIM\ \!\!$\uparrow$\!\!\!\!
\\ 
\midrule
\!\!\!VMPI\!\!\! & \!\!\!0.333\!\!\! & \!\!\!26.57\!\!\! & \!\!\!0.820\!\!\! & \!\!\!0.364\!\!\! & \!\!\!24.45\!\!\! & \!\!\!0.748\!\!\! & \!\!\!0.344\!\!\! & \!\!\!21.53\!\!\! & \!\!\!0.712\!\!\! & \!\!\!0.355\!\!\! & \!\!\!20.62\!\!\! & \!\!\!0.663\!\!\! \\
\!\!\!SVMPI++\!\!\! & \!\!\!0.094\!\!\! & \!\!\!32.32\!\!\! & \!\!\!0.946\!\!\! & \!\!\!0.137\!\!\! & \!\!\!28.43\!\!\! & \!\!\!0.892\!\!\! & \!\!\!0.162\!\!\! & \!\!\!23.68\!\!\! & \!\!\!0.799\!\!\! & \!\!\!0.178\!\!\! & \!\!\!\textbf{22.76}\!\!\! & \!\!\!0.744\!\!\! \\
\!\!\!MINE++\!\!\! & \!\!\!0.117\!\!\! & \!\!\!31.40\!\!\! & \!\!\!0.940\!\!\! & \!\!\!0.168\!\!\! & \!\!\!27.94\!\!\! & \!\!\!0.879\!\!\! & \!\!\!0.169\!\!\! & \!\!\!23.45\!\!\! & \!\!\!\textbf{0.802}\!\!\! & \!\!\!0.197\!\!\! & \!\!\!22.55\!\!\! & \!\!\!\textbf{0.749}\!\!\! \\
\!\!\!3D-Photo\!\!\! & \!\!\!0.069\!\!\! & \!\!\!30.28\!\!\! & \!\!\!0.909\!\!\! &  \!\!\!0.130\!\!\! & \!\!\!27.10\!\!\! & \!\!\!0.851\!\!\! & \!\!\!0.178\!\!\! & \!\!\!21.72\!\!\! & \!\!\!0.730\!\!\! & \!\!\!0.174\!\!\! & \!\!\!21.91\!\!\! & \!\!\!0.727\!\!\!\\
\!\!\!\emph{Ours}\!\!\! & \!\!\!\textbf{0.059}\!\!\! & \!\!\!\textbf{34.21}\!\!\! & \!\!\!\textbf{0.966}\!\!\! & \!\!\!\textbf{0.115}\!\!\! & \!\!\!\textbf{29.13}\!\!\! & \!\!\!\textbf{0.903}\!\!\! & \!\!\!\textbf{0.145}\!\!\! & \!\!\!\textbf{23.76}\!\!\! & \!\!\!\textbf{0.802}\!\!\! & \!\!\!\textbf{0.169}\!\!\! & \!\!\!22.71\!\!\! & \!\!\!0.742\!\!\! \\
% Ours-ft &  &  &  &  &  &  &  &  &  &  &  & \\
\bottomrule
\end{tabular}
\label{table:benchmark}
\vspace{-2pt}
\end{table}

\begin{table}
    \centering
    \scriptsize
    \caption{Ablation study on network design, regularization term, and training data.}
    \vspace{-8pt}
    \begin{tabular}{lccc}
    \toprule
    & LPIPS~$\downarrow$ & PSNR~$\uparrow$ & SSIM~$\uparrow$ \\
    \midrule
    w/o PAN & 0.080 & 32.38 & 0.935 \\
    w/o FM & 0.120 & 31.43 & 0.929 \\
    \midrule
    w/o $\mathcal{L}_{rank}$ & 0.079 & 32.26 & 0.932 \\
    w/o $\mathcal{L}_{assign}$ & 0.205 & 27.61 & 0.846 \\
    \midrule
    MiDaS \& hole & 0.075 & 32.61 & 0.937 \\
    DPT \& general & 0.076 & 32.74 & 0.940 \\
    \midrule
    \emph{Ours} & \textbf{0.073} & \textbf{32.76} & \textbf{0.942} \\
    \bottomrule
    \end{tabular}
    \label{table:ablate}
\end{table}

\subsection{Ablation Study}
We investigate the effect of network architecture, loss function, and training data, using the Ken Burns dataset and 16 planes.

\subsubsection{Network Architecture}
As shown in Table~\ref{table:ablate}, the performance drops significantly if we discard the Plane Adjustment Network (PAN) and use predefined plane depth (\emph{e.g.}, 0.4dB drop in PSNR), demonstrating the advantage of our learned depth adjustment. If the feature masking scheme (FM) in the Color Prediction Network is removed (we directly concatenate the positional encoding of the plane depth to the shared feature map as in MINE~\cite{mine2021}), the quality also degrades significantly (64\% LPIPS increase). The inter-plane interactions injected by our feature masking mechanism are crucial to our adaptive MPI with scene-specific depth. 

\subsubsection{Loss Function}
Here we verify the effectiveness of the regularization terms in our loss function. 
Table~\ref{table:ablate} demonstrates that they play an important role in training our method: removing either $\mathcal{L}_{rank}$ or $\mathcal{L}_{assign}$ results in degraded accuracy, especially the latter.

\subsubsection{Training Data}
We further train our method using image pairs generated from another two strategies: 1) MiDaS \& hole -- we replace DPT with MiDaS~\cite{Ranftl2020} and train the inpainting network as described in Section~\ref{Sec:Method:Data}, and 2) DPT \& general --  we adopt the DPT depth and directly use an inpainting network pretrained on Place2~\cite{zhou2017places} dataset.  The results in Table~\ref{table:ablate} show that using DPT-estimated depth leads to significantly better results than using MiDaS, indicating that the quality of our generated training data can benefit from better monocular depth estimation.  Table~\ref{table:ablate} also shows that using our inpainting network leads to better results 
compared to a general-purpose one trained with random masks.

\begin{figure}[t]
    \centering
    \includegraphics[width=1.0\columnwidth]{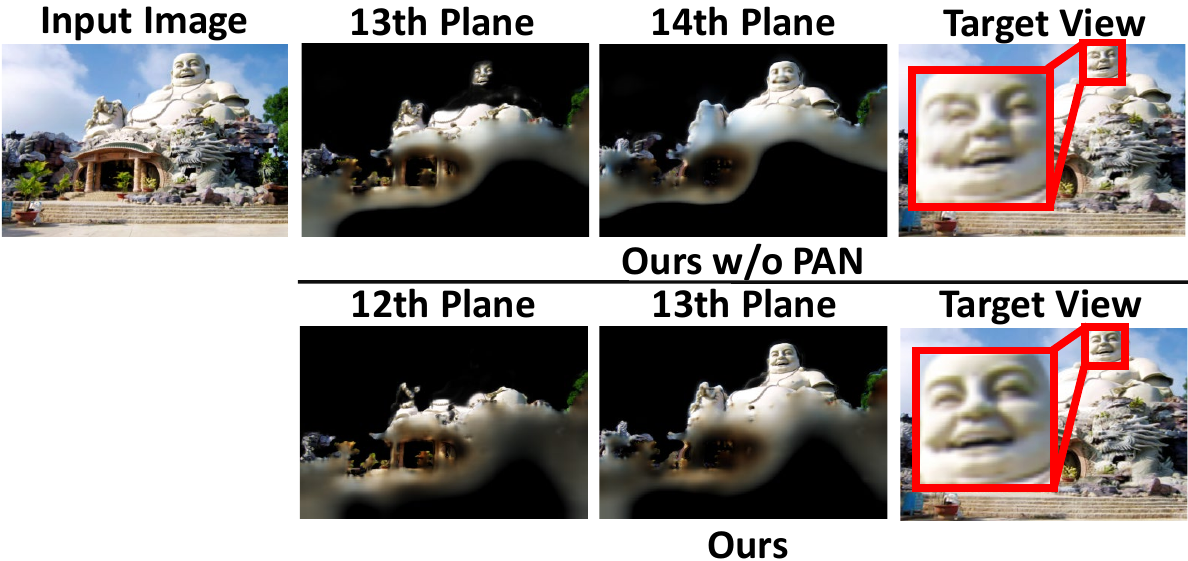}
    \vspace{-19pt}
    \caption{Visualization of alpha-multiplied color on MPIs predicted with and without plane adjustment. (\textbf{Best viewed with zoom-in.})}
    \label{Fig:visual_pan}
\end{figure}

\subsection{Analyzing the Plane Adjustment}
As shown in Fig.~\ref{Fig:visual_pan}, the scene-agnostic MPI (Ours w/o PAN) assigns the Buddha's face to two planes, leading to severe visual distortion in the novel view.
In contrast, our method learns to set one plane (the $13th$ one) to represent the Buddha's face according to the geometry and appearance information of the scene (see suppl. document for more visualization).
For reference, we report the average absolute corrections made by PAN in the normalized disparity space (ranging from 0 to 1) on the KenBurns dataset: 0.086, 0.058, 0.038, and 0.019 for 8, 16, 32, and 64 planes, respectively.

\subsection{Results on Higher Resolution Input}
We directly run our model (trained with $256\times 384$ images) on $512\times 768$ inputs, and found the model generalizes quite well to the higher resolution. The visual results are shown in the suppl. video.

\begin{figure}[t]
    \centering
    \includegraphics[width=1.0\columnwidth]{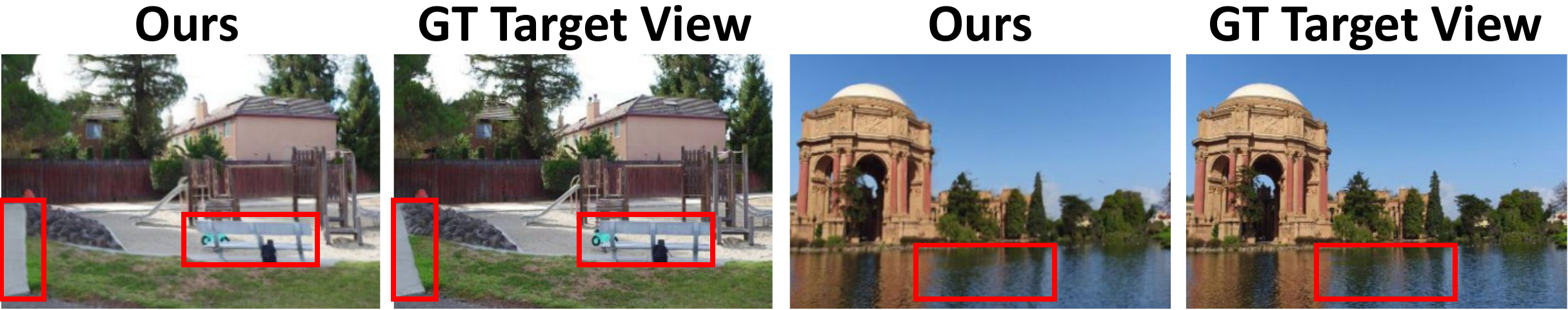}
    \vspace{-19pt}
    \caption{Limitations. Left: erroneous depth estimation of the ground and tree leads to visual distortions.
    Right: our method cannot model view-dependent effects such as reflection. See the supplementary video for more examples.}
    \label{Fig:failure}
\end{figure}

\subsection{Limitations}
\label{sec:limit}
Our method still has several limitations. 
As shown in Fig.~\ref{Fig:failure}, erroneous depth estimation leads to visual distortions in the synthesis results. Since our training data is generated by warping single-view images, the network cannot synthesize view-dependent effects such as reflection. 
Besides, our method tends to generate relatively blurry content or artifacts in occluded regions (\emph{e.g.}, see Fig.~\ref{Fig:visual_slide}).
Adding adversarial loss that focused on  occluded regions might be a possible remedy.
We currently use a relatively low resolution to train our method.
Collecting a large-scale high-resolution dataset and training our method on it is an interesting future direction.
Like other MPI-based methods, our method still has limited ability in  representing slanted surfaces, even though it can reduce the artifacts in this case by allocating more planes to model them.

\section{Conclusion}
We have presented a novel \emph{AdaMPI} architecture to deal with the challenging task of synthesizing novel views from single-view images in the wild.
The two key ingredients are a Plane Adjustment Network for MPI plane depth adjustment from an initial configuration and a Color Prediction Network for depth-aware color and density prediction with a novel feature masking scheme.
To train our method, a simple yet effective \emph{warp-back} strategy is proposed to obtain large-scale multi-view training data using only unconstrained single-view images.
Our method achieves state-of-the-art view synthesis results on both synthetic and real datasets.

{
\bibliographystyle{ACM-Reference-Format}
\bibliography{sample-manuscript}


\begin{thebibliography}{52}

%%% the Web address.

\ifx \showCODEN    \undefined \def \showCODEN     #1{\unskip}     \fi
\ifx \showDOI      \undefined \def \showDOI       #1{#1}\fi
\ifx \showISBNx    \undefined \def \showISBNx     #1{\unskip}     \fi
\ifx \showISBNxiii \undefined \def \showISBNxiii  #1{\unskip}     \fi
\ifx \showISSN     \undefined \def \showISSN      #1{\unskip}     \fi
\ifx \showLCCN     \undefined \def \showLCCN      #1{\unskip}     \fi
\ifx \shownote     \undefined \def \shownote      #1{#1}          \fi
\ifx \showarticletitle \undefined \def \showarticletitle #1{#1}   \fi
\ifx \showURL      \undefined \def \showURL       {\relax}        \fi
% invisible to TeX
\providecommand\bibfield[2]{#2}
\providecommand\bibinfo[2]{#2}
\providecommand\natexlab[1]{#1}
\providecommand\showeprint[2][]{arXiv:#2}

\bibitem[Aleotti et~al\mbox{.}(2021)]%
        {Aleotti_CVPR_2021}
\bibfield{author}{\bibinfo{person}{Filippo Aleotti}, \bibinfo{person}{Matteo
  Poggi}, {and} \bibinfo{person}{Stefano Mattoccia}.}
  \bibinfo{year}{2021}\natexlab{}.
\newblock \showarticletitle{Learning optical flow from still images}. In
  \bibinfo{booktitle}{\emph{CVPR}}.
\newblock


\bibitem[Aliev et~al\mbox{.}(2020)]%
        {aliev2020neural}
\bibfield{author}{\bibinfo{person}{Kara-Ali Aliev}, \bibinfo{person}{Artem
  Sevastopolsky}, \bibinfo{person}{Maria Kolos}, \bibinfo{person}{Dmitry
  Ulyanov}, {and} \bibinfo{person}{Victor Lempitsky}.}
  \bibinfo{year}{2020}\natexlab{}.
\newblock \showarticletitle{Neural point-based graphics}. In
  \bibinfo{booktitle}{\emph{ECCV}}. Springer, \bibinfo{pages}{696--712}.
\newblock


\bibitem[Caesar et~al\mbox{.}(2018)]%
        {coco2018cvpr}
\bibfield{author}{\bibinfo{person}{Holger Caesar}, \bibinfo{person}{Jasper
  Uijlings}, {and} \bibinfo{person}{Vittorio Ferrari}.}
  \bibinfo{year}{2018}\natexlab{}.
\newblock \showarticletitle{COCO-Stuff: Thing and stuff classes in context}. In
  \bibinfo{booktitle}{\emph{CVPR}}. IEEE.
\newblock


\bibitem[Chen and Koltun(2017)]%
        {chen2017photographic}
\bibfield{author}{\bibinfo{person}{Qifeng Chen} {and} \bibinfo{person}{Vladlen
  Koltun}.} \bibinfo{year}{2017}\natexlab{}.
\newblock \showarticletitle{Photographic image synthesis with cascaded
  refinement networks}. In \bibinfo{booktitle}{\emph{ICCV}}.
  \bibinfo{pages}{1511--1520}.
\newblock


\bibitem[Dai et~al\mbox{.}(2020)]%
        {dai2020neural}
\bibfield{author}{\bibinfo{person}{Peng Dai}, \bibinfo{person}{Yinda Zhang},
  \bibinfo{person}{Zhuwen Li}, \bibinfo{person}{Shuaicheng Liu}, {and}
  \bibinfo{person}{Bing Zeng}.} \bibinfo{year}{2020}\natexlab{}.
\newblock \showarticletitle{Neural point cloud rendering via multi-plane
  projection}. In \bibinfo{booktitle}{\emph{CVPR}}.
  \bibinfo{pages}{7830--7839}.
\newblock


\bibitem[Flynn et~al\mbox{.}(2019)]%
        {flynn2019deepview}
\bibfield{author}{\bibinfo{person}{John Flynn}, \bibinfo{person}{Michael
  Broxton}, \bibinfo{person}{Paul Debevec}, \bibinfo{person}{Matthew DuVall},
  \bibinfo{person}{Graham Fyffe}, \bibinfo{person}{Ryan Overbeck},
  \bibinfo{person}{Noah Snavely}, {and} \bibinfo{person}{Richard Tucker}.}
  \bibinfo{year}{2019}\natexlab{}.
\newblock \showarticletitle{Deepview: View synthesis with learned gradient
  descent}. In \bibinfo{booktitle}{\emph{CVPR}}. \bibinfo{pages}{2367--2376}.
\newblock


\bibitem[Godard et~al\mbox{.}(2019)]%
        {godard2019digging}
\bibfield{author}{\bibinfo{person}{Cl{\'e}ment Godard}, \bibinfo{person}{Oisin
  Mac~Aodha}, \bibinfo{person}{Michael Firman}, {and}
  \bibinfo{person}{Gabriel~J Brostow}.} \bibinfo{year}{2019}\natexlab{}.
\newblock \showarticletitle{Digging into self-supervised monocular depth
  estimation}. In \bibinfo{booktitle}{\emph{ICCV}}.
  \bibinfo{pages}{3828--3838}.
\newblock


\bibitem[Hartley and Zisserman(2004)]%
        {Hartley2004}
\bibfield{author}{\bibinfo{person}{R.~I. Hartley} {and} \bibinfo{person}{A.
  Zisserman}.} \bibinfo{year}{2004}\natexlab{}.
\newblock \bibinfo{booktitle}{\emph{Multiple View Geometry in Computer Vision}
  (\bibinfo{edition}{second} ed.)}.
\newblock \bibinfo{publisher}{Cambridge University Press, ISBN: 0521540518}.
\newblock


\bibitem[He et~al\mbox{.}(2016)]%
        {he2016deep}
\bibfield{author}{\bibinfo{person}{Kaiming He}, \bibinfo{person}{Xiangyu
  Zhang}, \bibinfo{person}{Shaoqing Ren}, {and} \bibinfo{person}{Jian Sun}.}
  \bibinfo{year}{2016}\natexlab{}.
\newblock \showarticletitle{Deep residual learning for image recognition}. In
  \bibinfo{booktitle}{\emph{CVPR}}. \bibinfo{pages}{770--778}.
\newblock


\bibitem[Hu et~al\mbox{.}(2021)]%
        {hu2021worldsheet}
\bibfield{author}{\bibinfo{person}{Ronghang Hu}, \bibinfo{person}{Nikhila
  Ravi}, \bibinfo{person}{Alex Berg}, {and} \bibinfo{person}{Deepak Pathak}.}
  \bibinfo{year}{2021}\natexlab{}.
\newblock \showarticletitle{Worldsheet: Wrapping the World in a 3D Sheet for
  View Synthesis from a Single Image}. In \bibinfo{booktitle}{\emph{ICCV}}.
\newblock


\bibitem[Im et~al\mbox{.}(2019)]%
        {im2019dpsnet}
\bibfield{author}{\bibinfo{person}{Sunghoon Im}, \bibinfo{person}{Hae-Gon
  Jeon}, \bibinfo{person}{Stephen Lin}, {and} \bibinfo{person}{In~So Kweon}.}
  \bibinfo{year}{2019}\natexlab{}.
\newblock \showarticletitle{Dpsnet: End-to-end deep plane sweep stereo}.
\newblock \bibinfo{journal}{\emph{ICLR}} (\bibinfo{year}{2019}).
\newblock


\bibitem[Jampani et~al\mbox{.}(2021)]%
        {jampani:ICCV:2021}
\bibfield{author}{\bibinfo{person}{Varun Jampani}, \bibinfo{person}{Huiwen
  Chang}, \bibinfo{person}{Kyle Sargent}, \bibinfo{person}{Abhishek Kar},
  \bibinfo{person}{Richard Rucker}, \bibinfo{person}{Michael Krainin},
  \bibinfo{person}{Dominik Kaeser}, \bibinfo{person}{William~T Freeman},
  \bibinfo{person}{David Salesin}, \bibinfo{person}{Brian Curless}, {and}
  \bibinfo{person}{Ce Liu}.} \bibinfo{year}{2021}\natexlab{}.
\newblock \showarticletitle{SLIDE: Single Image 3D Photography with Soft
  Layering and Depth-aware Inpainting}. In \bibinfo{booktitle}{\emph{ICCV}}.
\newblock


\bibitem[Jiang et~al\mbox{.}(2021)]%
        {jiang2021focal}
\bibfield{author}{\bibinfo{person}{Liming Jiang}, \bibinfo{person}{Bo Dai},
  \bibinfo{person}{Wayne Wu}, {and} \bibinfo{person}{Chen~Change Loy}.}
  \bibinfo{year}{2021}\natexlab{}.
\newblock \showarticletitle{Focal Frequency Loss for Image Reconstruction and
  Synthesis}. In \bibinfo{booktitle}{\emph{ICCV}}.
\newblock


\bibitem[Kellnhofer et~al\mbox{.}(2021)]%
        {Kellnhofer:2021:nlr}
\bibfield{author}{\bibinfo{person}{Petr Kellnhofer}, \bibinfo{person}{Lars
  Jebe}, \bibinfo{person}{Andrew Jones}, \bibinfo{person}{Ryan Spicer},
  \bibinfo{person}{Kari Pulli}, {and} \bibinfo{person}{Gordon Wetzstein}.}
  \bibinfo{year}{2021}\natexlab{}.
\newblock \showarticletitle{Neural Lumigraph Rendering}. In
  \bibinfo{booktitle}{\emph{CVPR}}.
\newblock


\bibitem[Kingma and Ba(2014)]%
        {kingma2014adam}
\bibfield{author}{\bibinfo{person}{Diederik~P Kingma} {and}
  \bibinfo{person}{Jimmy Ba}.} \bibinfo{year}{2014}\natexlab{}.
\newblock \showarticletitle{Adam: A method for stochastic optimization}.
\newblock \bibinfo{journal}{\emph{arXiv preprint arXiv:1412.6980}}
  (\bibinfo{year}{2014}).
\newblock


\bibitem[Knapitsch et~al\mbox{.}(2017)]%
        {Knapitsch2017}
\bibfield{author}{\bibinfo{person}{Arno Knapitsch}, \bibinfo{person}{Jaesik
  Park}, \bibinfo{person}{Qian-Yi Zhou}, {and} \bibinfo{person}{Vladlen
  Koltun}.} \bibinfo{year}{2017}\natexlab{}.
\newblock \showarticletitle{Tanks and Temples: Benchmarking Large-Scale Scene
  Reconstruction}.
\newblock \bibinfo{journal}{\emph{ACM Transactions on Graphics}}
  \bibinfo{volume}{36}, \bibinfo{number}{4} (\bibinfo{year}{2017}).
\newblock


\bibitem[Kopf et~al\mbox{.}(2019)]%
        {kopf2019practical}
\bibfield{author}{\bibinfo{person}{Johannes Kopf}, \bibinfo{person}{Suhib
  Alsisan}, \bibinfo{person}{Francis Ge}, \bibinfo{person}{Yangming Chong},
  \bibinfo{person}{Kevin Matzen}, \bibinfo{person}{Ocean Quigley},
  \bibinfo{person}{Josh Patterson}, \bibinfo{person}{Jossie Tirado},
  \bibinfo{person}{Shu Wu}, {and} \bibinfo{person}{Michael~F Cohen}.}
  \bibinfo{year}{2019}\natexlab{}.
\newblock \showarticletitle{Practical 3D photography}. In
  \bibinfo{booktitle}{\emph{CVPR Workshops}}.
\newblock


\bibitem[Kopf et~al\mbox{.}(2020)]%
        {kopf2020one}
\bibfield{author}{\bibinfo{person}{Johannes Kopf}, \bibinfo{person}{Kevin
  Matzen}, \bibinfo{person}{Suhib Alsisan}, \bibinfo{person}{Ocean Quigley},
  \bibinfo{person}{Francis Ge}, \bibinfo{person}{Yangming Chong},
  \bibinfo{person}{Josh Patterson}, \bibinfo{person}{Jan-Michael Frahm},
  \bibinfo{person}{Shu Wu}, \bibinfo{person}{Matthew Yu}, {et~al\mbox{.}}}
  \bibinfo{year}{2020}\natexlab{}.
\newblock \showarticletitle{One shot 3d photography}.
\newblock \bibinfo{journal}{\emph{ACM Transactions on Graphics (TOG)}}
  \bibinfo{volume}{39}, \bibinfo{number}{4} (\bibinfo{year}{2020}),
  \bibinfo{pages}{76--1}.
\newblock


\bibitem[Lai et~al\mbox{.}(2021)]%
        {lai2021video}
\bibfield{author}{\bibinfo{person}{Zihang Lai}, \bibinfo{person}{Sifei Liu},
  \bibinfo{person}{Alexei~A Efros}, {and} \bibinfo{person}{Xiaolong Wang}.}
  \bibinfo{year}{2021}\natexlab{}.
\newblock \showarticletitle{Video Autoencoder: self-supervised disentanglement
  of static 3D structure and motion}. In \bibinfo{booktitle}{\emph{ICCV}}.
  \bibinfo{pages}{9730--9740}.
\newblock


\bibitem[Li et~al\mbox{.}(2021)]%
        {mine2021}
\bibfield{author}{\bibinfo{person}{Jiaxin Li}, \bibinfo{person}{Zijian Feng},
  \bibinfo{person}{Qi She}, \bibinfo{person}{Henghui Ding},
  \bibinfo{person}{Changhu Wang}, {and} \bibinfo{person}{Gim~Hee Lee}.}
  \bibinfo{year}{2021}\natexlab{}.
\newblock \showarticletitle{MINE: Towards Continuous Depth MPI with NeRF for
  Novel View Synthesis}. In \bibinfo{booktitle}{\emph{ICCV}}.
\newblock


\bibitem[Li and Khademi~Kalantari(2020)]%
        {Li2020LF}
\bibfield{author}{\bibinfo{person}{Qinbo Li} {and} \bibinfo{person}{Nima
  Khademi~Kalantari}.} \bibinfo{year}{2020}\natexlab{}.
\newblock \showarticletitle{Synthesizing Light Field From a Single Image with
  Variable MPI and Two Network Fusion}.
\newblock \bibinfo{journal}{\emph{ACM Transactions on Graphics}}
  \bibinfo{volume}{39}, \bibinfo{number}{6}.
\newblock
\urldef\tempurl%
\url{https://doi.org/10.1145/3414685.3417785}
\showDOI{\tempurl}


\bibitem[Liu et~al\mbox{.}(2020)]%
        {liu2020neural}
\bibfield{author}{\bibinfo{person}{Lingjie Liu}, \bibinfo{person}{Jiatao Gu},
  \bibinfo{person}{Kyaw~Zaw Lin}, \bibinfo{person}{Tat-Seng Chua}, {and}
  \bibinfo{person}{Christian Theobalt}.} \bibinfo{year}{2020}\natexlab{}.
\newblock \showarticletitle{Neural Sparse Voxel Fields}.
\newblock \bibinfo{journal}{\emph{NIPS}} (\bibinfo{year}{2020}).
\newblock


\bibitem[Lombardi et~al\mbox{.}(2019)]%
        {lombardi2019neural}
\bibfield{author}{\bibinfo{person}{Stephen Lombardi}, \bibinfo{person}{Tomas
  Simon}, \bibinfo{person}{Jason Saragih}, \bibinfo{person}{Gabriel Schwartz},
  \bibinfo{person}{Andreas Lehrmann}, {and} \bibinfo{person}{Yaser Sheikh}.}
  \bibinfo{year}{2019}\natexlab{}.
\newblock \showarticletitle{Neural volumes: Learning dynamic renderable volumes
  from images}.
\newblock \bibinfo{journal}{\emph{SIGGRAPH}} (\bibinfo{year}{2019}).
\newblock


\bibitem[Luvizon et~al\mbox{.}(2021)]%
        {luvizon2021adaptive}
\bibfield{author}{\bibinfo{person}{Diogo~C Luvizon}, \bibinfo{person}{Gustavo
  Sutter~P Carvalho}, \bibinfo{person}{Andreza~A dos Santos},
  \bibinfo{person}{Jhonatas~S Conceicao}, \bibinfo{person}{Jose~L
  Flores-Campana}, \bibinfo{person}{Luis~GL Decker}, \bibinfo{person}{Marcos~R
  Souza}, \bibinfo{person}{Helio Pedrini}, \bibinfo{person}{Antonio Joia},
  {and} \bibinfo{person}{Otavio~AB Penatti}.} \bibinfo{year}{2021}\natexlab{}.
\newblock \showarticletitle{Adaptive multiplane image generation from a single
  internet picture}. In \bibinfo{booktitle}{\emph{Proceedings of the IEEE/CVF
  Winter Conference on Applications of Computer Vision}}.
  \bibinfo{pages}{2556--2565}.
\newblock


\bibitem[Mildenhall et~al\mbox{.}(2019)]%
        {mildenhall2019local}
\bibfield{author}{\bibinfo{person}{Ben Mildenhall}, \bibinfo{person}{Pratul~P
  Srinivasan}, \bibinfo{person}{Rodrigo Ortiz-Cayon},
  \bibinfo{person}{Nima~Khademi Kalantari}, \bibinfo{person}{Ravi Ramamoorthi},
  \bibinfo{person}{Ren Ng}, {and} \bibinfo{person}{Abhishek Kar}.}
  \bibinfo{year}{2019}\natexlab{}.
\newblock \showarticletitle{Local light field fusion: Practical view synthesis
  with prescriptive sampling guidelines}.
\newblock \bibinfo{journal}{\emph{ACM Transactions on Graphics (TOG)}}
  \bibinfo{volume}{38}, \bibinfo{number}{4} (\bibinfo{year}{2019}),
  \bibinfo{pages}{1--14}.
\newblock


\bibitem[Mildenhall et~al\mbox{.}(2020)]%
        {mildenhall2020nerf}
\bibfield{author}{\bibinfo{person}{Ben Mildenhall}, \bibinfo{person}{Pratul~P
  Srinivasan}, \bibinfo{person}{Matthew Tancik}, \bibinfo{person}{Jonathan~T
  Barron}, \bibinfo{person}{Ravi Ramamoorthi}, {and} \bibinfo{person}{Ren Ng}.}
  \bibinfo{year}{2020}\natexlab{}.
\newblock \showarticletitle{Nerf: Representing scenes as neural radiance fields
  for view synthesis}. In \bibinfo{booktitle}{\emph{ECCV}}. Springer,
  \bibinfo{pages}{405--421}.
\newblock


\bibitem[Nazeri et~al\mbox{.}(2019)]%
        {Nazeri_2019_ICCV}
\bibfield{author}{\bibinfo{person}{Kamyar Nazeri}, \bibinfo{person}{Eric Ng},
  \bibinfo{person}{Tony Joseph}, \bibinfo{person}{Faisal Qureshi}, {and}
  \bibinfo{person}{Mehran Ebrahimi}.} \bibinfo{year}{2019}\natexlab{}.
\newblock \showarticletitle{EdgeConnect: Structure Guided Image Inpainting
  using Edge Prediction}. In \bibinfo{booktitle}{\emph{ICCV Workshops}}.
\newblock


\bibitem[Nguyen-Phuoc et~al\mbox{.}(2019)]%
        {nguyen2019hologan}
\bibfield{author}{\bibinfo{person}{Thu Nguyen-Phuoc}, \bibinfo{person}{Chuan
  Li}, \bibinfo{person}{Lucas Theis}, \bibinfo{person}{Christian Richardt},
  {and} \bibinfo{person}{Yong-Liang Yang}.} \bibinfo{year}{2019}\natexlab{}.
\newblock \showarticletitle{Hologan: Unsupervised learning of 3d
  representations from natural images}. In \bibinfo{booktitle}{\emph{ICCV}}.
  \bibinfo{pages}{7588--7597}.
\newblock


\bibitem[Niklaus et~al\mbox{.}(2019)]%
        {Niklaus_TOG_2019}
\bibfield{author}{\bibinfo{person}{Simon Niklaus}, \bibinfo{person}{Long Mai},
  \bibinfo{person}{Jimei Yang}, {and} \bibinfo{person}{Feng Liu}.}
  \bibinfo{year}{2019}\natexlab{}.
\newblock \showarticletitle{3D Ken Burns Effect from a Single Image}.
\newblock \bibinfo{journal}{\emph{ACM Transactions on Graphics}}
  \bibinfo{volume}{38}, \bibinfo{number}{6} (\bibinfo{year}{2019}),
  \bibinfo{pages}{184:1--184:15}.
\newblock


\bibitem[Porter and Duff(1984)]%
        {alpha_comp}
\bibfield{author}{\bibinfo{person}{Thomas Porter} {and} \bibinfo{person}{Tom
  Duff}.} \bibinfo{year}{1984}\natexlab{}.
\newblock \showarticletitle{Compositing Digital Images}. In
  \bibinfo{booktitle}{\emph{Proceedings of the 11th Annual Conference on
  Computer Graphics and Interactive Techniques}}
  \emph{(\bibinfo{series}{SIGGRAPH '84})}. \bibinfo{publisher}{Association for
  Computing Machinery}, \bibinfo{address}{New York, NY, USA},
  \bibinfo{pages}{253–259}.
\newblock
\showISBNx{0897911385}
\urldef\tempurl%
\url{https://doi.org/10.1145/800031.808606}
\showDOI{\tempurl}


\bibitem[Ranftl et~al\mbox{.}(2021)]%
        {Ranftl2021}
\bibfield{author}{\bibinfo{person}{Ren\'{e} Ranftl}, \bibinfo{person}{Alexey
  Bochkovskiy}, {and} \bibinfo{person}{Vladlen Koltun}.}
  \bibinfo{year}{2021}\natexlab{}.
\newblock \showarticletitle{Vision Transformers for Dense Prediction}.
\newblock \bibinfo{journal}{\emph{ICCV}} (\bibinfo{year}{2021}).
\newblock


\bibitem[Ranftl et~al\mbox{.}(2020)]%
        {Ranftl2020}
\bibfield{author}{\bibinfo{person}{Ren\'{e} Ranftl}, \bibinfo{person}{Katrin
  Lasinger}, \bibinfo{person}{David Hafner}, \bibinfo{person}{Konrad
  Schindler}, {and} \bibinfo{person}{Vladlen Koltun}.}
  \bibinfo{year}{2020}\natexlab{}.
\newblock \showarticletitle{Towards Robust Monocular Depth Estimation: Mixing
  Datasets for Zero-shot Cross-dataset Transfer}.
\newblock \bibinfo{journal}{\emph{IEEE Transactions on Pattern Analysis and
  Machine Intelligence (TPAMI)}} (\bibinfo{year}{2020}).
\newblock


\bibitem[Riegler and Koltun(2021)]%
        {Riegler2021SVS}
\bibfield{author}{\bibinfo{person}{Gernot Riegler} {and}
  \bibinfo{person}{Vladlen Koltun}.} \bibinfo{year}{2021}\natexlab{}.
\newblock \showarticletitle{Stable View Synthesis}. In
  \bibinfo{booktitle}{\emph{CVPR}}.
\newblock


\bibitem[Rockwell et~al\mbox{.}(2021)]%
        {Rockwell2021}
\bibfield{author}{\bibinfo{person}{Chris Rockwell}, \bibinfo{person}{David~F.
  Fouhey}, {and} \bibinfo{person}{Justin Johnson}.}
  \bibinfo{year}{2021}\natexlab{}.
\newblock \showarticletitle{PixelSynth: Generating a 3D-Consistent Experience
  from a Single Image}. In \bibinfo{booktitle}{\emph{ICCV}}.
\newblock


\bibitem[Rombach et~al\mbox{.}(2021)]%
        {rombach2021geometry}
\bibfield{author}{\bibinfo{person}{Robin Rombach}, \bibinfo{person}{Patrick
  Esser}, {and} \bibinfo{person}{Bj{\"o}rn Ommer}.}
  \bibinfo{year}{2021}\natexlab{}.
\newblock \showarticletitle{Geometry-Free View Synthesis: Transformers and no
  3D Priors}.
\newblock \bibinfo{journal}{\emph{ICCV}} (\bibinfo{year}{2021}).
\newblock


\bibitem[Ronneberger et~al\mbox{.}(2015)]%
        {ronneberger2015u}
\bibfield{author}{\bibinfo{person}{Olaf Ronneberger}, \bibinfo{person}{Philipp
  Fischer}, {and} \bibinfo{person}{Thomas Brox}.}
  \bibinfo{year}{2015}\natexlab{}.
\newblock \showarticletitle{U-net: Convolutional networks for biomedical image
  segmentation}. In \bibinfo{booktitle}{\emph{MICCAI}}. Springer,
  \bibinfo{pages}{234--241}.
\newblock


\bibitem[Sch\"{o}nberger and Frahm(2016)]%
        {schoenberger2016sfm}
\bibfield{author}{\bibinfo{person}{Johannes~Lutz Sch\"{o}nberger} {and}
  \bibinfo{person}{Jan-Michael Frahm}.} \bibinfo{year}{2016}\natexlab{}.
\newblock \showarticletitle{Structure-from-Motion Revisited}. In
  \bibinfo{booktitle}{\emph{CVPR}}.
\newblock


\bibitem[Shih et~al\mbox{.}(2020)]%
        {Shih3DP20}
\bibfield{author}{\bibinfo{person}{Meng-Li Shih}, \bibinfo{person}{Shih-Yang
  Su}, \bibinfo{person}{Johannes Kopf}, {and} \bibinfo{person}{Jia-Bin Huang}.}
  \bibinfo{year}{2020}\natexlab{}.
\newblock \showarticletitle{3D Photography using Context-aware Layered Depth
  Inpainting}. In \bibinfo{booktitle}{\emph{CVPR}}.
\newblock


\bibitem[Srinivasan et~al\mbox{.}(2019)]%
        {srinivasan2019pushingmpi}
\bibfield{author}{\bibinfo{person}{Pratul~P Srinivasan},
  \bibinfo{person}{Richard Tucker}, \bibinfo{person}{Jonathan~T Barron},
  \bibinfo{person}{Ravi Ramamoorthi}, \bibinfo{person}{Ren Ng}, {and}
  \bibinfo{person}{Noah Snavely}.} \bibinfo{year}{2019}\natexlab{}.
\newblock \showarticletitle{Pushing the boundaries of view extrapolation with
  multiplane images}. In \bibinfo{booktitle}{\emph{CVPR}}.
  \bibinfo{pages}{175--184}.
\newblock


\bibitem[Srinivasan et~al\mbox{.}(2017)]%
        {srinivasan2017learning}
\bibfield{author}{\bibinfo{person}{Pratul~P Srinivasan},
  \bibinfo{person}{Tongzhou Wang}, \bibinfo{person}{Ashwin Sreelal},
  \bibinfo{person}{Ravi Ramamoorthi}, {and} \bibinfo{person}{Ren Ng}.}
  \bibinfo{year}{2017}\natexlab{}.
\newblock \showarticletitle{Learning to synthesize a 4D RGBD light field from a
  single image}. In \bibinfo{booktitle}{\emph{ICCV}}.
  \bibinfo{pages}{2243--2251}.
\newblock


\bibitem[Tucker and Snavely(2020)]%
        {single_view_mpi}
\bibfield{author}{\bibinfo{person}{Richard Tucker} {and} \bibinfo{person}{Noah
  Snavely}.} \bibinfo{year}{2020}\natexlab{}.
\newblock \showarticletitle{Single-view View Synthesis with Multiplane Images}.
  In \bibinfo{booktitle}{\emph{CVPR}}.
\newblock


\bibitem[Tulsiani et~al\mbox{.}(2018)]%
        {tulsiani2018layer}
\bibfield{author}{\bibinfo{person}{Shubham Tulsiani}, \bibinfo{person}{Richard
  Tucker}, {and} \bibinfo{person}{Noah Snavely}.}
  \bibinfo{year}{2018}\natexlab{}.
\newblock \showarticletitle{Layer-structured 3d scene inference via view
  synthesis}. In \bibinfo{booktitle}{\emph{ECCV}}. \bibinfo{pages}{302--317}.
\newblock


\bibitem[Vaswani et~al\mbox{.}(2017)]%
        {vaswani2017attention}
\bibfield{author}{\bibinfo{person}{Ashish Vaswani}, \bibinfo{person}{Noam
  Shazeer}, \bibinfo{person}{Niki Parmar}, \bibinfo{person}{Jakob Uszkoreit},
  \bibinfo{person}{Llion Jones}, \bibinfo{person}{Aidan~N Gomez},
  \bibinfo{person}{{\L}ukasz Kaiser}, {and} \bibinfo{person}{Illia
  Polosukhin}.} \bibinfo{year}{2017}\natexlab{}.
\newblock \showarticletitle{Attention is all you need}. In
  \bibinfo{booktitle}{\emph{NIPS}}. \bibinfo{pages}{5998--6008}.
\newblock


\bibitem[Wang et~al\mbox{.}(2021)]%
        {wang2021ibrnet}
\bibfield{author}{\bibinfo{person}{Qianqian Wang}, \bibinfo{person}{Zhicheng
  Wang}, \bibinfo{person}{Kyle Genova}, \bibinfo{person}{Pratul~P Srinivasan},
  \bibinfo{person}{Howard Zhou}, \bibinfo{person}{Jonathan~T Barron},
  \bibinfo{person}{Ricardo Martin-Brualla}, \bibinfo{person}{Noah Snavely},
  {and} \bibinfo{person}{Thomas Funkhouser}.} \bibinfo{year}{2021}\natexlab{}.
\newblock \showarticletitle{Ibrnet: Learning multi-view image-based rendering}.
  In \bibinfo{booktitle}{\emph{CVPR}}. \bibinfo{pages}{4690--4699}.
\newblock


\bibitem[Wang et~al\mbox{.}(2020)]%
        {wang2020tartanair}
\bibfield{author}{\bibinfo{person}{Wenshan Wang}, \bibinfo{person}{Delong Zhu},
  \bibinfo{person}{Xiangwei Wang}, \bibinfo{person}{Yaoyu Hu},
  \bibinfo{person}{Yuheng Qiu}, \bibinfo{person}{Chen Wang},
  \bibinfo{person}{Yafei Hu}, \bibinfo{person}{Ashish Kapoor}, {and}
  \bibinfo{person}{Sebastian Scherer}.} \bibinfo{year}{2020}\natexlab{}.
\newblock \showarticletitle{Tartanair: A dataset to push the limits of visual
  slam}. In \bibinfo{booktitle}{\emph{IROS}}. IEEE,
  \bibinfo{pages}{4909--4916}.
\newblock


\bibitem[Watson et~al\mbox{.}(2020)]%
        {watson-2020-stereo-from-mono}
\bibfield{author}{\bibinfo{person}{Jamie Watson}, \bibinfo{person}{Oisin~Mac
  Aodha}, \bibinfo{person}{Daniyar Turmukhambetov}, \bibinfo{person}{Gabriel~J.
  Brostow}, {and} \bibinfo{person}{Michael Firman}.}
  \bibinfo{year}{2020}\natexlab{}.
\newblock \showarticletitle{Learning Stereo from Single Images}. In
  \bibinfo{booktitle}{\emph{ECCV}}.
\newblock


\bibitem[Wiles et~al\mbox{.}(2020)]%
        {wiles2020synsin}
\bibfield{author}{\bibinfo{person}{Olivia Wiles}, \bibinfo{person}{Georgia
  Gkioxari}, \bibinfo{person}{Richard Szeliski}, {and} \bibinfo{person}{Justin
  Johnson}.} \bibinfo{year}{2020}\natexlab{}.
\newblock \showarticletitle{{SynSin}: {E}nd-to-end View Synthesis from a Single
  Image}. In \bibinfo{booktitle}{\emph{CVPR}}.
\newblock


\bibitem[Yu et~al\mbox{.}(2021)]%
        {yu2021pixelnerf}
\bibfield{author}{\bibinfo{person}{Alex Yu}, \bibinfo{person}{Vickie Ye},
  \bibinfo{person}{Matthew Tancik}, {and} \bibinfo{person}{Angjoo Kanazawa}.}
  \bibinfo{year}{2021}\natexlab{}.
\newblock \showarticletitle{pixelnerf: Neural radiance fields from one or few
  images}. In \bibinfo{booktitle}{\emph{CVPR}}. \bibinfo{pages}{4578--4587}.
\newblock


\bibitem[Yu et~al\mbox{.}(2019)]%
        {yu2019free}
\bibfield{author}{\bibinfo{person}{Jiahui Yu}, \bibinfo{person}{Zhe Lin},
  \bibinfo{person}{Jimei Yang}, \bibinfo{person}{Xiaohui Shen},
  \bibinfo{person}{Xin Lu}, {and} \bibinfo{person}{Thomas~S Huang}.}
  \bibinfo{year}{2019}\natexlab{}.
\newblock \showarticletitle{Free-form image inpainting with gated convolution}.
  In \bibinfo{booktitle}{\emph{ICCV}}. \bibinfo{pages}{4471--4480}.
\newblock


\bibitem[Zhang et~al\mbox{.}(2018)]%
        {zhang2018perceptual}
\bibfield{author}{\bibinfo{person}{Richard Zhang}, \bibinfo{person}{Phillip
  Isola}, \bibinfo{person}{Alexei~A Efros}, \bibinfo{person}{Eli Shechtman},
  {and} \bibinfo{person}{Oliver Wang}.} \bibinfo{year}{2018}\natexlab{}.
\newblock \showarticletitle{The Unreasonable Effectiveness of Deep Features as
  a Perceptual Metric}. In \bibinfo{booktitle}{\emph{CVPR}}.
\newblock


\bibitem[Zhou et~al\mbox{.}(2017)]%
        {zhou2017places}
\bibfield{author}{\bibinfo{person}{Bolei Zhou}, \bibinfo{person}{Agata
  Lapedriza}, \bibinfo{person}{Aditya Khosla}, \bibinfo{person}{Aude Oliva},
  {and} \bibinfo{person}{Antonio Torralba}.} \bibinfo{year}{2017}\natexlab{}.
\newblock \showarticletitle{Places: A 10 million Image Database for Scene
  Recognition}.
\newblock \bibinfo{journal}{\emph{TPAMI}} (\bibinfo{year}{2017}).
\newblock


\bibitem[Zhou et~al\mbox{.}(2018)]%
        {zhou2018stereo}
\bibfield{author}{\bibinfo{person}{Tinghui Zhou}, \bibinfo{person}{Richard
  Tucker}, \bibinfo{person}{John Flynn}, \bibinfo{person}{Graham Fyffe}, {and}
  \bibinfo{person}{Noah Snavely}.} \bibinfo{year}{2018}\natexlab{}.
\newblock \showarticletitle{Stereo Magnification: Learning View Synthesis using
  Multiplane Images}. In \bibinfo{booktitle}{\emph{SIGGRAPH}}.
\newblock


\end{thebibliography}
}
\clearpage

\renewcommand{\thesection}{\Alph{section}}
\renewcommand{\thefigure}{\Roman{figure}}
\renewcommand{\thetable}{\Roman{table}}
\renewcommand{\theequation}{\Roman{equation}}
\setcounter{figure}{0}
\setcounter{equation}{0}
\setcounter{section}{0}

\begin{minipage}[t]{1.0\textwidth}
		\centering
		{\huge{\emph{\textbf{Supplementary Material}}}} \\
		\vspace{15pt}
		\includegraphics[width=1.0\linewidth]{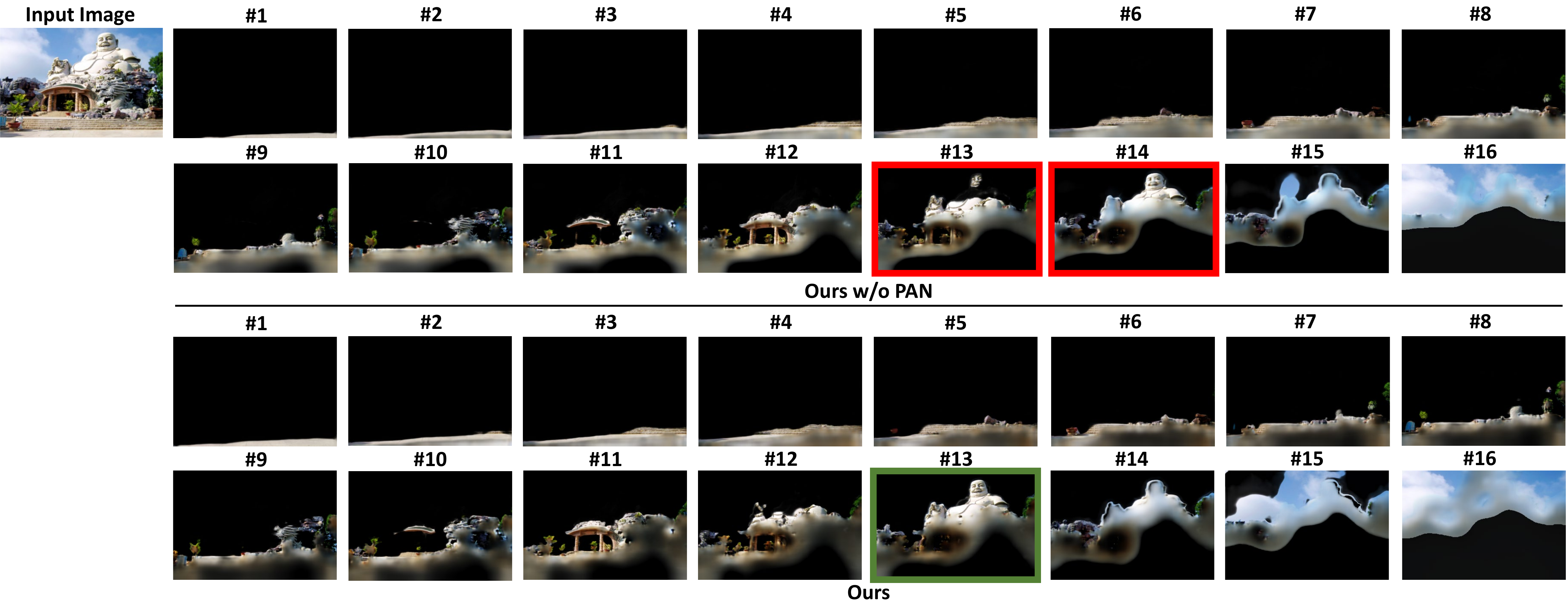}
		\vspace{-15pt}
		\captionof{figure}{Visualization of alpha-multiplied color planes with and without plane adjustment.}
		\label{Fig:vis_pan}
		\vspace{18pt}
\end{minipage}

\begin{figure*}[t]
	\centering
	\includegraphics[width=1.0\linewidth]{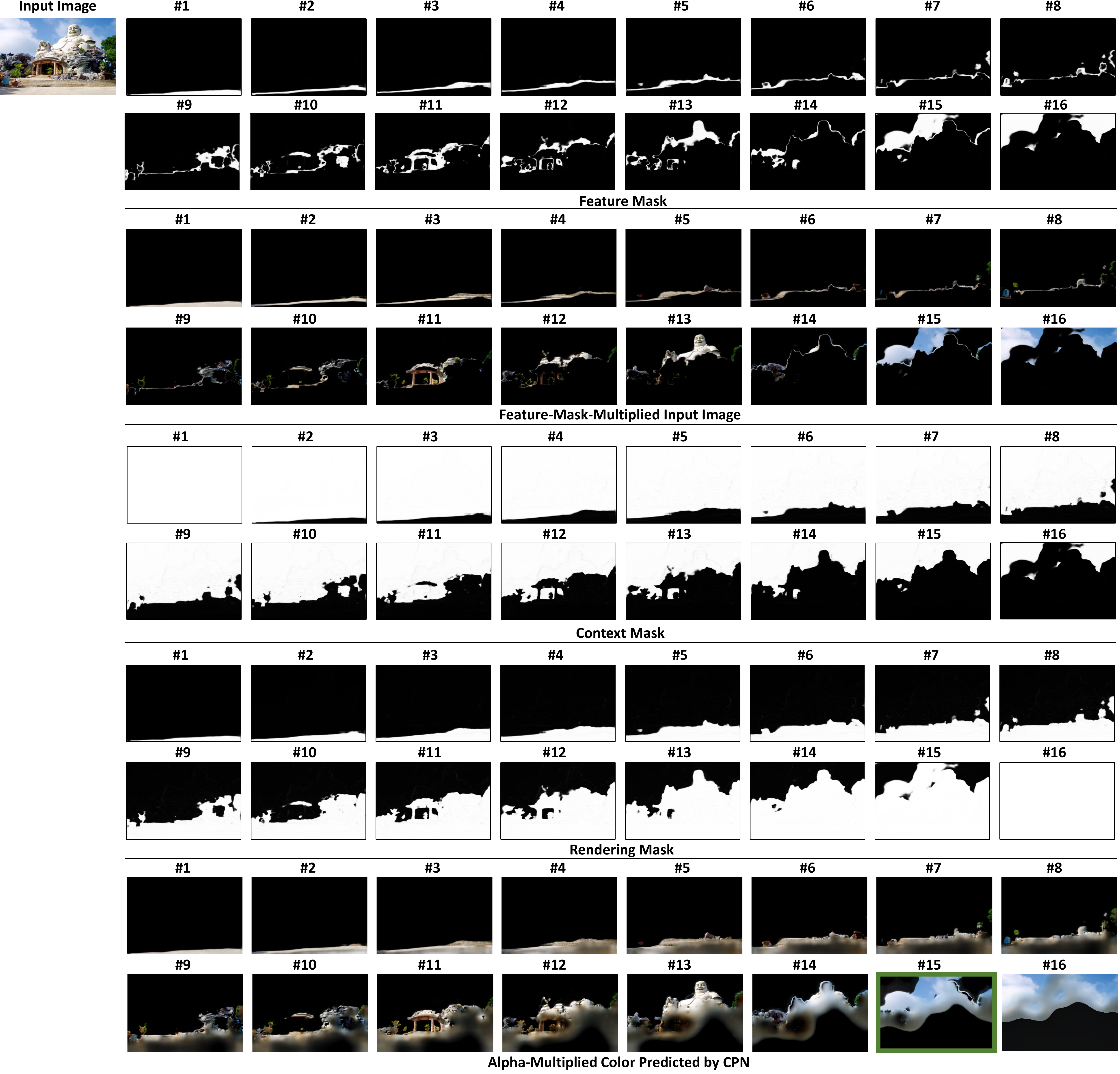}
	\caption{Visualization of the color planes and the feature masks.}
	\label{Fig:vis_cpn}
\end{figure*}

\begin{figure*}[t]
	\centering
	\includegraphics[width=1.0\linewidth]{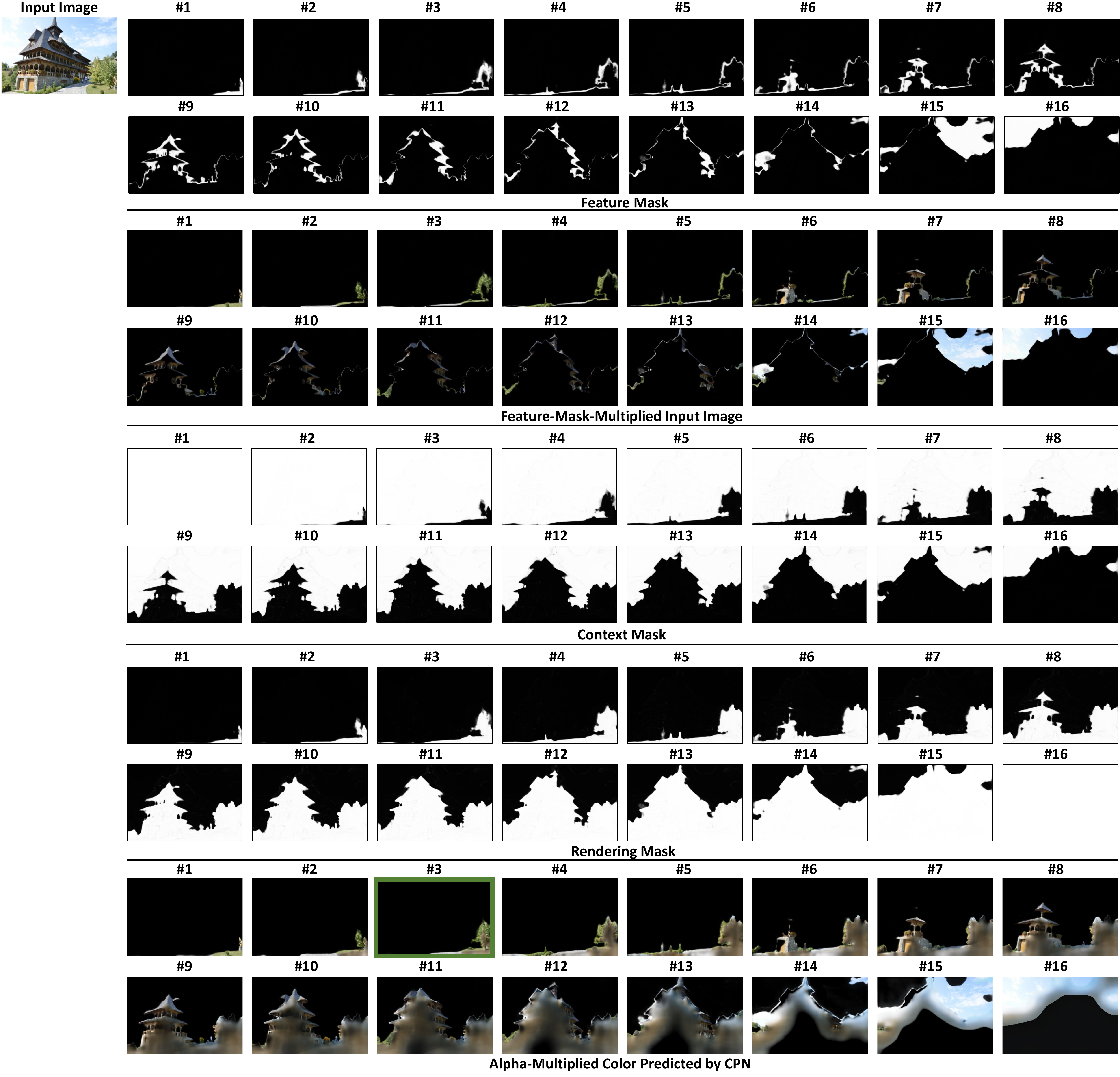}
	\caption{Visualization of the color planes and the feature masks.}
	\label{Fig:vis_cpn_2}
\end{figure*}

\begin{minipage}[t]{1.0\textwidth}
\begin{multicols}{2}
\section{Test Data Selection and Processing}
We use four datasets for method evaluation which provide multi-view images or videos of static scenes: Ken Burns~\cite{Niklaus_TOG_2019}, TartanAir~\cite{wang2020tartanair}, RealEstate-10K~\cite{zhou2018stereo}, and Tank \& Temples~\cite{Knapitsch2017}. 
\emph{Ken Burns} and \emph{TartanAir} are synthetic datasets rendered with graphics engine, which provide ground truth depth maps and camera parameters.
We randomly sampled 1.5K and 260 stereo pairs from these two datasets for testing. 
\emph{RealEstate-10K} 
contains video clips of real-world indoor scenes.
We sampled 800 clips from the test set for evaluation. 
For each clip, we specify a randomly sampled frame as the source view and the following $10$th frame as the target view.
Following~\cite{Shih3DP20}, we use DPSNet~\cite{im2019dpsnet} to estimate depth and use the camera parameters estimated by COLMAP~\cite{schoenberger2016sfm}.
\emph{Tank \& Temples} contains multi-view images of real in-the-wild scenes.
We use the depth maps estimated by DPT and align its scale with the camera parameters recovered by COLMAP. 1K images are sampled from this dataset for evaluation.
We select image pairs with moderate camera motion to evaluate our method.

\section{Runtime}
Our method with 64 planes takes $0.072$s, including $0.004$s for depth adjustment and $0.068$s for color prediction, to generate the MPI for a $256\times 384$ image on a Nvidia Tesla V100 GPU.
For comparison, the VMPI method takes $0.003$s, SVMPI takes $0.013$s, MINE takes $0.011$s, and our closest competitor in quality, 3D-Photo, takes several seconds.

\section{Visualization of our Learned Planes}
In Fig.~\ref{Fig:vis_pan}, we show all the color planes predicted with or without the Plane Adjustment Network, for the case we presented in Section 4.4 and Figure 7 in the main paper.
We observe that the scene-agnostic MPI (Ours w/o PAN) assigns the Buddha's face to two planes (the $13 th$ and $14 th$ one, highlighted with the red box), while our method learns to set one plane (the $13 th$ one, highlighted with the green box) to represent it, leading to better view synthesis results with less visual distortion.

\noindent\quad Figure~\ref{Fig:vis_cpn} and \ref{Fig:vis_cpn_2} further show the masks and color predicted by our method on each plane (16 planes are used in these cases). Guided by the feature masking scheme, our Color Prediction Network (CPN) learns to represent the visible pixels on the planes and inpaint the occluded content especially near occlusion boundary.
We highlight two typical samples with the green box: on the $15 th$ plane in Fig.~\ref{Fig:vis_cpn_2}, the sky region occluded by the Buddha's face are inpainted;
on the $3 rd$ plane in Fig.~\ref{Fig:vis_cpn}, CPN learns to extend the grass texture to inner regions.
\end{multicols}
\end{minipage}

\end{document}